\documentclass[review]{elsarticle}

\usepackage{hyperref}
\usepackage[utf8]{inputenc}
\usepackage[T1]{fontenc}

\journal{Neurocomputing}
\usepackage{setspace}
\usepackage{graphicx}
\graphicspath{{figures/}}
\usepackage{subfig}
\usepackage[table]{xcolor}
\usepackage{multirow}
\usepackage{booktabs}
\usepackage{array}
\usepackage{xcolor, soul}
\definecolor{LightGreen}{HTML}{B6FF00}
\sethlcolor{LightGreen}
\usepackage{url}
\newcommand{\quotes}[1]{``#1''}
\usepackage{amsmath,amssymb,bm}
\usepackage[ruled,vlined]{algorithm2e}

\begin{document}

\begin{frontmatter}

\title{An Adaptive Locally Connected Neuron Model: Focusing Neuron}

\author{F. Boray Tek \corref{mycorrespondingauthor}}
\address{Department of Computer Engineering, Isik University, Sile, Istanbul, Turkey, 34980}
\cortext[mycorrespondingauthor]{Corresponding author}
\ead{boray.tek@isikun.edu.tr}


\begin{abstract}
This paper presents a new artificial neuron model capable of learning its receptive field in the topological domain of inputs. 
The model provides adaptive and differentiable local connectivity (plasticity) applicable to any domain. It requires no other tool than the backpropagation algorithm to learn its parameters which control the receptive field locations and apertures. This research explores whether this ability makes the neuron focus on informative inputs and yields any advantage over fully connected neurons.
The experiments include tests of focusing neuron networks of one or two hidden layers on synthetic and well-known image recognition data sets. The results demonstrated that the focusing neurons can move their receptive fields towards more informative inputs. In the simple two-hidden layer networks, the focusing layers outperformed the dense layers in the classification of the 2D spatial data sets. Moreover, the focusing networks performed better than the dense networks even when 70$\%$ of the weights were pruned. The tests on convolutional networks revealed that using focusing layers instead of dense layers for the classification of convolutional features may work better in some data sets.
\end{abstract}


\end{frontmatter}


\section{Introduction}
The structure of the brain is mutable. Neuroplasticity, the ability of neurons to reorganize and adapt to inner or outer environments, causes the brain to change through its lifetime \citep{charles2009,merzenich2014}. The changes can occur in larger (e.g., cortical remapping) or in microscopic synaptic scales in which individual cells alter their connections through activity and learning \citep{merzenich2014,power2010}. The changes allow divisions and specializations to make the biological brain capable of solving thousands of sensory, cognitive, and behavioral problems within the same framework \citep{menon_2015}. 

Recent developments in the design and training of artificial neural networks (ANN) enabled us to solve many pattern recognition problems within an acceptable range of accuracy. However, artificial networks are not yet comparable to the self-organizing and multi-functional biological brain. Many biology-inspired artificial models are unfit for large-scale pattern recognition tasks \citep{bartunov2018}. Hence, still, many recent works propose sophisticated artificial network structures optimized for solving single-target tasks \citep{szegedy2016, larsson2017, urban2017, srivastava2015, xu2015, ba2014}. Although some architectures can deal with multimodal information \citep{xu2015, Vinyals2015}, the typical design approach is to create a fixed topology (network connection map) to solve a task. The connection map is pre-determined by an expert, who uses heuristics or prior knowledge of the domain. Some hard-wired networks are composed of several interacting sub-networks \citep{larsson2017}; however, the fixed network structures still lack the self-organizing capacity of the brain.

Mimicking the self-organizing brain may be possible with an automated algorithm that can construct a network iteratively or via evolutionary mechanisms \cite{Floreano2008, soltoggio_2018}, yet new models that can self-create a topological structure are nevertheless required. Though the literature contains examples of architecture optimizers \citep{romero2015,baker2017,liu2018,coates2011}, network growing/pruning algorithms \citep{fiesler1994,hassibi93,han2015,cortes2017}, and evolutionary processes \citep{soltoggio_2018}, the approach of the current paper is based on a new artificial neuron type that can learn its receptive field and thereby its local connections in a topological structure. The new model, termed a \quotes{focusing neuron}, brings an adaptive-wiring ability to artificial neurons.

Some studies on local learning or local receptive fields \cite{Serre_2007,Masquelier_2007,olshausen_1996} interpret the locality of a neuron as its selectivity in the input value domain. For example, in an input space formed by two input features ($x$,$y$), such a neuron learns to be active for only a subset of the input space (e.g., $\left[ x_a<x<x_b,\, y_c<y<y_d \right]$). Hence, this localization corresponds to a partitioning (clustering) of the input (value) space but requires the neuron to be globally connected to observe the whole space. Instead, the locality of the proposed model here is based on the input topology.

A focusing neuron has a focus attachment to change its local receptive field in the topological domain of inputs \citep{cam2017, tek_2019}. It can learn and create unique connection maps for inputs and problems presented by data. Akin to the synaptic plasticity guided by biochemical cues between axons and dendritic spines \citep{stoeckli2017,tracey2017}, the new model uses error gradients to guide the receptive field. To that end, the proposed framework assumes a continuous positional (spatial) space for the inputs in which the focus function is differentiable. Thus, focusing neurons are made entirely trainable using the gradient descent algorithm with no additional heuristics.

To summarize, this paper introduces a topology-aware and locally adaptive neuron model, the focusing neuron. It does not claim the state-of-the-art in a particular pattern recognition application, nor does it suggest replacing the fully connected neuron model or convolutional networks. However, the new model and its components (positional inputs and the training of the locality) can guide us in reducing structural heuristics and redundancy in neural networks.

The main contributions made by this paper are as follows:
\begin{enumerate}
	\item It formally describes the focusing neuron model.
	\item It devises a scheme for the initialization of weights and focus parameters.
	\item It demonstrates that the focusing neuron can seek informative features and avoid redundant inputs.
	\item It presents test results from networks of focusing layers in comparison with those from networks of fully connected layers, using the popular MNIST \citep{lecun1998a}, MNIST-cluttered \citep{jaderberg2015}, Fashion \citep{xiao2017}, CIFAR-10 \citep{cifar10}, LFW-Faces \citep{lfw}, Reuters newswires \citep{keras}, DNA \cite{OpenML2013}, and Boston Housing data sets \citep{keras}.
	\item It demonstrates that, owing to their (trained) narrower receptive fields, connections constructed by a focusing neuron layer can be $70\%$ sparser than those of its dense counterpart, while also performing better.
	\item It demonstrates that when used as a classifier after stacked convolutional layers, focusing layers can provide a better or similar classification performance compared to dense layers.
\end{enumerate}

\section{Background}

The concept of locality is important for neural networks and the motivation of this work. This section reviews the related literature and examines the fully and locally connected neuron models from the corresponding perspective.

\subsection{Local Learning} \label{localcon}
Hubel and Wiesel \citep{Hubel_1962} discovered that the \emph{simple cells} in the primary visual cortex are selective for the position, scale, orientation, and polarity of inputs. Further, the \emph{complex cells} that process signals from these simple cells can be selective in as much as to activate only in the presence of a particular face or object \citep{Chang_2017}. The selectivity (or locality) of a biological neuron thus relies on different cell types, local wiring (selective pooling), and the hierarchical structure of the neural circuitry \citep{Poggio_2013}. 

A psychologist, Donald Hebbian \citep{Hebb}, was the first to refer to a metabolic mechanism that strengthens the connection between two neurons: \quotes{When an axon of cell A
	is near enough to excite cell B and repeatedly or persistently takes
	part in firing it, some growth process or metabolic change takes
	place in one or both cells such that A's efficiency, as one of the
	cells firing B, is increased}. 
The description is slightly vague. However, note that the axon of A (not A itself) is required to be sufficiently \emph{close} enough to strengthen the connection with neuron B. 

From the AI perspective, Baldi and Sadowski's framework \citep{baldi_2016} maintains that 1) local learning depends on local information (of pre- and post-synaptic neurons), and 2) the functional form which operates the learning rule must use only the local information due to the activations of these neurons. Thus, for a generic update rule of the form, $\bigtriangleup w = f(x,\Theta)$ the input signal $x$ and update parameters $\Theta$ must both be local.

\subsubsection{Local Receptive Fields}
The discussion of local receptive fields can be traced back to Rosenblatt's \citep{Rosenblatt} perceptron, which was the first \emph{weighted} neuron model devised for feed-forward networks. Some variations of the perceptron were designed to be locally connected, i.e., had a limited \emph{fan-in} in the input space \citep{Minsky_1969}. The perceptron also included random (stochastic), partially connected neurons. As Minsky \citep{Minsky_1969} had also noticed, non-overlapping local receptive fields limit the (spatial) scale of information that can be extracted from the input. Fukushima \citep{fukushima_1983} subsequently tackled this problem with a hierarchical network structure (aka neocognitron), which later inspired convolutional neural networks \citep{lecun1998a}. 

\subsection{Fully Connected Neuron}
The conventional fully connected neuron model \citep{haykin1998,hagan2014} is often used in feedforward neural networks. The neuron multiplies the inputs ($x_1,x_2,...x_m$) by the connection weights ($w_1,w_2,...w_m$) and then sums the weighted inputs, adding a bias term ($b$) to calculate a net/total input. It then passes the net input ($net$) through a non-linear transfer function ($f$) to produce the output ($a$) of the neuron (\ref{ffeq}).
\begin{gather}
a =  f\:\Bigg( \sum_{i=1}^{m} w_{i}x_{i} + b \Bigg)
\label{ffeq}
\end{gather} 

The receptive (or input) field covers every input or neuron in the backward connected layer and is thus defined as being \quotes{fully connected}. A layer formed of such neurons is commonly referred to as a \quotes{dense layer}. For a fully connected neuron of $m$ inputs, there can be 2$^m$ permutations (of the input), which are equivalent since the positioning (ordering) of the inputs (here the index $i$) is unimportant. The free parameters of the model are the weight values (and the bias) which are updated (trained) using the delta rule (\ref{delta}):
\begin{gather}
\hat{w_i}= w_i-\eta\frac{\partial E}{\partial w_i} = w_i -\eta\frac{\partial E}{\partial a}\frac{\partial a}{\partial net}\frac{\partial net}{\partial w_i} \\ \hat{w_i}= w_i -\eta\frac{\partial E}{\partial a}f^{'}(net)x_i
\label{delta}
\end{gather} 

\subsubsection{Redundant Connections}
At first glance, a fully connected neuron should be able to zero-out some of its weights through updates and thence become locally, or at least partially connected. However, no matter how redundant the inputs or connections, the training cannot cancel any connection, partly due to the learning rule and the ordinary cost functions such as minimum squared error or cross-entropy.

To understand this phenomenon, let us examine the delta rule for a neuron with two inputs and two connections. Assume that one connection ($w_1$) is attached to a constant, non-informative, and non-zero input ($x_1\!=\!c$), where the other connection is attached to an informative and changing input ($x_2$). Since $x_1$ is constant, let us set the corresponding weight to zero, i.e., $w_1=0$, which results in a broken connection. In the first (next) update, $w_1$ cannot remain at zero unless either the error term $\frac{\partial E}{\partial a}$ or the derivative term $f^{'}(net)$ is also zero ($\eta>0$). Both terms are shared in the update of both weights ($w_1, w_2$). Hence, the error term cannot be zero unless the training is completed. The derivative term depends on the activation function ($f$) in the form of sigmoid, tanh, or relu, which have positive (or zero) derivative values. A zero derivative value can keep the weight $w_1$ at zero, yet this prevents the update of $w_2$ as well. Even if the weight value reaches zero via an update, it would change at the next update unless the error term had also reached zero in the same iteration. The induction is similar for a neuron with a greater number of inputs or a truly random (non-informative) redundant input. I have prepared a python notebook to demonstrate this phenomenon experimentally on the MNIST data set \citep{gitcode}.

The problem of network redundancy in terms of the number of neurons and connections was a popular theme of early studies \citep{hassibi93, lecun1990,elizondo1997}. Subsequent increases in computational power or excitement around novel, deeper architectures may have temporarily overshadowed the problem. However, research into the issue of redundancy has recently regained its momentum with the development of resource-critical applications \citep{coates2011,howard2017,han2015deep,Manessi2017,wu2016quantized}.

It is possible to reduce network redundancy through the simple strategy of adding a regularization factor to the loss calculation. Typical examples are the L1 or L2 norms of weights in a network or layer \citep{goodfellow2016}. Such a regularization term creates a penalty for the weight magnitudes, pushing them toward zero, and ultimately making the network sparser and less redundant. However, the penalty term is applied to all connection weights in a layer (or whole network); it does not explicitly target redundant connections. Therefore, adding a regularization penalty to a cost function requires attention because it may compete with the target loss of a network.

\subsubsection{Locally Connected Neuron Model}
A locally and partially connected model may refer to various structures \citep{elizondo1997,kung1988,lecun_1989_local}. \emph{Local} implies a connected and bounded group (subset) of connections, whereas partial may refer to any subset of all input connections \citep{elizondo1997}. For example, connecting a random subset of neurons will create a partially connected structure. Therefore, partial includes local, but not vice versa. 

The fully connected model (\ref{ffeq}) can express a partially connected neuron by setting some weights at zero. However, it neither generates nor maintains a partially connected structure during the training due to the gradient updates deriving from the backpropagation learning rule.  Therefore, to create locally (and partially) connected structures in practice, some designers have used domain knowledge or assumptions to partition the input space and create manual wirings for specific problems \citep{taigman2014, rowley1998, gregor2010, munder_2006}. The fixed local connection maps (or layers) constrain the neurons to receive signals from a fixed subset of inputs, i.e., a region in the input domain. In contrast, the proposed model allows a neuron to learn its local connections by trainable parameters.

\subsubsection{Convolutional Layers}
Several attempts to create adaptive local connections for convolutional neurons are reported in the literature. In locally smoothed neural networks \citep{pang2017}, convolutional filter weights were factorized into smoothing and kernel parts. The smoother part was modeled by a 2-D Gaussian function, and a regression network dynamically generated the parameters of the Gaussian smoothers. The Gaussian smoothing of the input space can be seen as a particular case of (dynamic) focusing neurons for two-dimensional input spaces. In \citep{cam_2018}, a Gaussian envelope was used to guide convolution kernels to create an adaptive receptive field size and orientation.

\subsubsection{Clustering based locality}
Further studies have implemented the locality of a neuron as its selectivity (specialization) in the input value domain \citep{Serre_2007,Masquelier_2007,olshausen_1996,rbf}. The selectivity is built by partitioning/clustering the input (feature) density space. A well-known example of clustering-based approaches is the radial basis function (RBF) network \citep{rbf}. In an RBF network, each neuron localizes itself in the input value domain using the center and spread of Gaussian kernels. RBF neurons cluster or partition the input value domain instead of the inputs' spatial domain. Likewise, learning in vector quantization networks involves competition between individual neurons to gain control of a cluster in the input domain \citep{Kohonen_lvq_1995}. RBFs (and LVQ neurons) are invariant to input topology, meaning that an RBF neuron must receive all inputs to localize itself in the input value domain.

A self-organizing map (SOM) projects the input feature space to a two-dimensional lattice \citep{kohonen1990}. Hence, SOM neurons can be described as neurons with spatial dimensions; however, the spatial dimension of inputs is ignored.

In a hierarchical network structure, the neurons are activated selectively for different inputs. For example, in the primate brain, one can find individual neurons activating for particular faces \citep{Chang_2017}. The neural hierarchy naturally induces a specialization (localization) in the input space. Some research aims at manipulating or guiding this selectivity (and specialization) to create orthogonal and sparser receptive fields and thence representations \citep{Serre_2007,Masquelier_2007,olshausen_1996}.


\subsubsection{Adaptive locality in the topological space}
This section describes how locality involves local connections/wiring of the neurons, as in our model. The biological name for adaptive local connectivity is plasticity (neuroplasticity). While relatively less-explored in the artificial domain, plasticity has long been an active area of neurobiology research \citep{esposito_2016}, as well as in spiking artificial neuron models \citep{bodenhausen_1990, Gerstner_2002, triesch_2007}. Nevertheless, some studies in artificial neural networks have directly aimed to reproduce Hebb’s rule or other plasticity models \citep{Oja1982,miconi_2018,huang_2015}. For instance, Miconi's \citep{miconi_2018} neuron model includes a fixed weight $w_{ij}$ (as in a fully connected neuron), a plasticity coefficient $\alpha_{i,j}$ and a recursively calculated Hebbian trace (Hebb$_{ij}$):
\begin{gather}
a_j(t) =  f\left(  \sum_{i=0}^N-1\left[ w_{ij} x_{ij} + \alpha_{i,j} \text{Hebb}_{ij}(t) a_i(t-1) \right]  \right)  
\label{fHebb}
\end{gather}
Thus, the model requires a recurrent calculation of Hebbian $\text{Hebb}_{ij}(t)$ trace using Oja's rule \citep{Oja1982}; however, the non-plastic weight component must be set at zero $w_{ij}=0$ to ensure complete plasticity.

Locality has been frequently investigated in the context of visual attention models where the objective is to interpret or exploit the information in various locations of an input image \citep{titi1998, Olshausen4700}. Recently proposed deep neural network architectures of visual attention models \citep{ba2014,xu2015} have shown impressive results in multiple object recognition, image captioning and similar tasks. The common strategy in these models is to process an input image in $n$ steps where, at each step, the recurrent layer examines the current input and then decides on the next location to process. 

Cheung et al.\ \citep{CheungWO16} proposed a retinal glimpse model (for visual attention) employing Gaussian kernels to control sampling locations and scales. Similarly, the focusing neuron model introduced in this paper employs Gaussian kernels to control and learn the location and scale of the local receptive field. The focusing neuron model is generic and hence may be used for any task without being limited to visual attention. 

The capsule architecture proposed by Sabour et al.~\citep{sabour2017} performs dynamic routing, which is an example of adaptive locality. A capsule is a group of neurons with a routing function that selectively and jointly directs the output vector of the group to only one of the several groups in the forward layer. The routing function is iteratively calculated per input. In contrast, the focusing model is static (learned during training) however, it can be made dynamic (calculated per input). Interestingly, the focusing neuron model can be made forward-faced, thus enabling it to adaptively focus the output transmitting field for the next layer. However, the backward focusing model is more compatible with the backpropagation algorithm and runs more efficiently on current graph computation frameworks such as Theano \citep{theano} or Tensorflow \citep{tensorflow}. 

\begin{figure}
	\centering
	\includegraphics[width=0.5\linewidth]{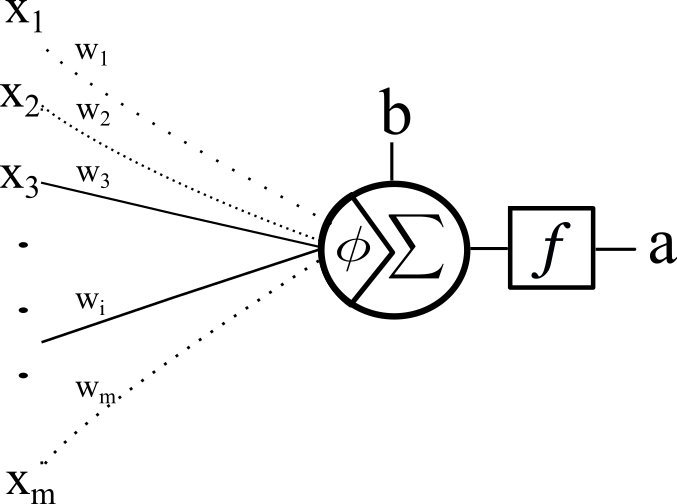}
	\caption{The focusing neuron model. The focus attachment $\phi$ allows the neuron to change its receptive field and adjust the aperture size. It generates coefficients which are multiplied by inputs and connection weights to collect the net input. The sum of net input and bias $b$ is mapped to the output by the activation transfer function $f$.}
	\label{fig:focus_neuron}
\end{figure}

\section{Method}

\subsection{Focusing neuron}

A focusing neuron is an adaptive, locally connected neuron. Figure~\ref{fig:focus_neuron} depicts the new model as a neuron with a backward focus attachment. The role of the focus is to suppress the connection weights for some inputs to allow signals from the rest. The name is motivated by the fact that the neuron can change both its local receptive field location and its size (aperture). The new model assumes a topological (positional) continuous input space where the focus function is differentiable.

The following describes a focus mechanism for a one-dimensional input topology which can be simply generalized to further dimensions. Let $\tau(i)\in\Re$ denote a mapping for the position of the input indexed with $i$. The focusing model has the following functional form:
\begin{equation}
a =  f\:\Bigg( \sum_{i=1}^{m} w_{i}\phi\left( \tau\left(i\right),\bm{\Theta\right)} x_{i} + b \Bigg) 
\label{ffneq}
\end{equation}

The focus function $\phi$ with its parameters  ($\bm{\Theta}$) generates a focus coefficient for each input $i$ at position $\tau(i)$. This may resemble the addition of a second weight for the connection. However, the coefficient values are dependent and controlled by the neuron's $\bm{\Theta}$ value. With the appropriate functional form and parameters, the neuron can change its input subset by moving or resizing its receptive aperture. This model complies with the popular tensor processing libraries which perform layer-wise operations. The effective (product) weight matrix $\bm{K}$ is computed by an element-wise multiplication of the focus coefficients $\bm{\Phi}$ with weights $\bm{W}$ (see Algorithm \ref{algo:1}).

\begin{algorithm}[H]
	\DontPrintSemicolon
	\KwIn{$\bm{X}$: $b \times m$ input matrix (b: batch, m: features), $\bm{\Phi}$: $m\times n$ (n: neurons) focus coefficent matrix, $\bm{W}$: $m\times n$ weight matrix, $\bm{b}$: $n\times 1$ bias vector (broadcasted into $b\times n$), $f$: activation function}
   \KwOut{$\bm{Y}$: $b \times n$ layer output.}
    \emph{$\circ$: element-wise Hadamard product; $.$: tensor dot}\; 
	\Begin{
		$\bm{K} = \bm{\Phi} \circ \bm{W}$\;		
		$\bm{Y} = f\left( \bm{X} . \bm{K} + \bm{b}\right) $\;		
	}
	\caption{Compute the output of a focused layer}
	\label{algo:1}
\end{algorithm}

\subsubsection{Focus control function}

A Gaussian form is the first candidate for the focus control function because it is continuous and differentiable, and it neither creates nor enhances extrema \citep{Lindeberg2011}. Some of these properties also exist in discrete space if the sample size is sufficiently large. For simplicity, it can be assumed that $\tau(i)=i/m$, so that the input position is taken as its normalized index over $m$ neurons in the input layer. In a multi-dimensional topology, both the control function and position are multivalued, e.g., $\tau(i,j)=(i/m, j/n)$. Nevertheless, a Gaussian focus $\phi(i,\bm{\Theta)}$ for single-dimensional input topologies can be defined in the following way:
\begin{equation}
\phi(i, \mu, \sigma) = s\: e ^{ - \frac{(i/m-\mu)^2}{2\sigma^2}}
\label{focus}
\end{equation}
where $\mu$ represents the center of the field and $\sigma$ acts as the aperture control (\ref{focus}). The form includes the scaler ($s$) in order to equalize the norm of $\phi$ to the norm of a fully connected equivalent model. The scaler $s$ maintains the norm with respect to the neuron's receptive field (\ref{norm}). When $\sigma$ is very large, the scaler converges to 1.0 ($s\approxeq1$), which makes the neuron fully connected.

\begin{equation}
s^j= \frac{\sqrt{m}}{\sqrt{\sum\limits_{i=1}^m \left(e ^{ - \frac{(i/m-\mu_j)^2}{2\sigma_j^2}}\right)^2}}
\label{norm}
\end{equation}

Algorithm \ref{algo:2} shows the procedure for computing Gaussian focus coefficients for all neurons in a layer. The procedure must be invoked prior to each calculation of the layer output in order to obtain the layer coefficient matrix.

\begin{algorithm}
	\DontPrintSemicolon
	\KwIn{Layer with $m$ inputs and $n$ neurons, $\bm{\mu}$: vector of $n$ focus centers,\\
		$\bm{\sigma}$: vector of $n$ focus apertures, 
		$\bm{\tau}$: vector of $m$ constant input locations (the paper experiments used $m$ linearly spaced points in range $[0.0, 1.0]$).}
	\KwOut{$\mathbf{\Phi}$: $m\times n$ focus coefficent matrix}
	\Begin{
		
		\For{$j\leftarrow 0$ \KwTo $n-1$}{
			$sum$ $\leftarrow 0$\\
			\For{$i\leftarrow 0$ \KwTo $m-1$}{
				$d \leftarrow \left(\bm{\tau}\! \left[i\right]-\bm{\mu}\!\left[j\right]\right) ^2/( 2\bm{\sigma}\!\left[j\right]^2)$ 
				
				$\bm{\Phi}\!\left[i,j\right] \leftarrow  e ^{-d} $
				
				$sum$ $\leftarrow $ $sum$ + $\bm{\Phi}\!\left[i,j\right]$\\
			}
			
			\emph{//Normalize the coefficients}\;
			\For{$i\leftarrow 0$ \KwTo $m-1$}{
				$\bm{\Phi}\!\left[i,j\right] \leftarrow \sqrt{m} \times \bm{\Phi}\!\left[i,j\right]/sum $\\
			}
		}
	}
	\caption{Compute the Gaussian Focus Coefficient Matrix of a Layer}
	\label{algo:2}
\end{algorithm}

Note that the Gaussian is nearly non-zero over the input positional domain. This is desirable since the neuron must receive inputs and corresponding gradient cues (however small) from the positions that are farther from the focus center. However, at run-time, the out-of-focus coefficients can be neglected (or pruned) if desired, which is different from pruning the smaller weights of a trained network. Since the product of the trained focus coefficients and weights form the effective weights, the focus function is not necessary at run-time, unless online learning is in progress.

The partial derivatives of the Gaussian focus function for $\mu$ and $\sigma$ are well defined (\ref{partial}). They can therefore into the chain rule easily (\ref{defoc}) and can be trained with the backpropagation learning rule and common cost functions:

\begin{equation}
\frac{\partial \phi}{\partial \mu} = \frac{(i/m-\mu)}{\sigma^2}\phi \; , \quad
\frac{\partial \phi}{\partial \sigma} = \frac{(i/m-\mu)^2}{\sigma^3}\phi
\label{partial}
\end{equation}

\begin{equation}
\hat{\mu}= \mu - \eta\frac{\partial E}{\partial a} \frac{\partial a}{\partial f}\frac{\partial f}{\partial \phi}\frac{\partial \phi}{\partial \mu} \; , \quad
\hat{\sigma}= \sigma - \eta\frac{\partial E}{\partial a} \frac{\partial a}{\partial f}\frac{\partial f}{\partial \phi}\frac{\partial \phi}{\partial \sigma} 
\label{defoc}
\end{equation}

There can be many alternatives to the Gaussian focus. A simple negation of Gaussian (i.e., $1-\phi$) or the derivative of Gaussian is a good candidate. However, one can design different focus control functions using various differentiable forms. For example, one can sample focus coefficients from a distribution such as Bernoulli or use a piece-wise linear function in the shape of a triangle.

\subsubsection{Initialization of parameters}
Recent studies have shown that the initialization of weights in a neural network is crucial for improving its training and generalization capacity \citep{he2015, glorot2010}. The focus coefficients scale the weights and change the variance of the propagated signals (see \ref{apx_w_init}). Moreover, since the total fan in (or the number of inputs) of a focusing neuron is usually larger than its effective fan in, a scaler (\ref{norm}) ensures the total norm to be equivalent to the norm of a fully connected neuron. However, in general, the weights of an individual focusing neuron ($j$) can be sampled with respect to the squared norm of its focus coefficients vector:
\begin{equation}
w^0\thicksim U\left[-\frac{\sqrt{6}}{\sqrt{\sum_{i=1}^{m} \phi^2(i)}}, \frac{\sqrt{6}}{\sqrt{\sum_{i=1}^{m} \phi^2(i)}} \right]
\label{init_sum}
\end{equation}

To initialize the receptive fields, one must also initialize $\mu_j$ and $\sigma_j$ for the Gaussian control function $\phi(i/m, (\mu_j, \sigma_j))$. The centers of the foci $\mu_j$ can be spread out or randomly initialized in the range $\left[0,1\right]$, while the aperture controls must be initialized to positive non-zero values, $\sigma_j>\epsilon$. Distributing the foci centers works more effectively than initializing them all in the center. Initializing $\sigma$ to a value in the range ($\left[0.02,0.2\right]$) enables the neuron to receive farther inputs. A smaller $\sigma$ value generates overly narrow apertures which may cause neurons to be stick in their initial position. On the other hand, a larger $\sigma$ generates a wider aperture which may cause the neurons to be indifferent and fully connected. In fact, the ideal $\mu$ and $\sigma$ initializations depend on the intended role of the neuron. In addition, one can apply L1 or L2 regularization on $\sigma$ to encourage locality, if the $\sigma\!=\!0$ case is handled explicitly.

\section{Experiments}

The experiments address two fundamental questions about the new model. First, can focusing neurons steer away from redundant inputs towards informative inputs? Second, how do focusing neurons perform in comparison to fully connected neurons? The code is shared online \citep{gitcode}.

\subsection{Learning to Focus}
First, a synthetic (Gaussian blob) classification data set of two classes and 20 dimensions was constructed. Then 20 additional Gaussian random input columns of normal distribution $N(0,1.0)$ were sampled and added to the left side or both sides of the data to form two separate data sets of 40 dimensions. Next, each data column was normalized for zero-mean and unit variance. 

\begin{figure}
	\centering
	\subfloat[]{\includegraphics[width=0.4\linewidth]{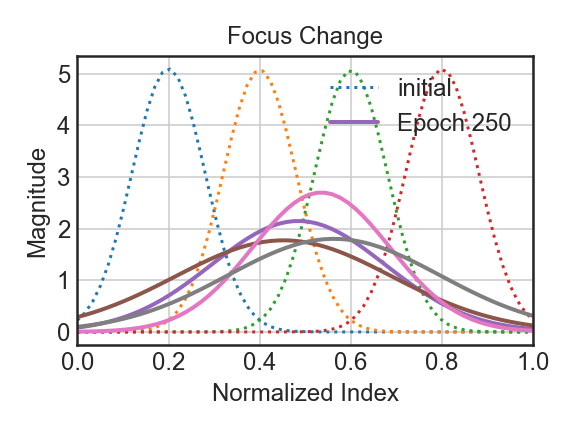}}	\subfloat[]{\includegraphics[width=0.4\linewidth]{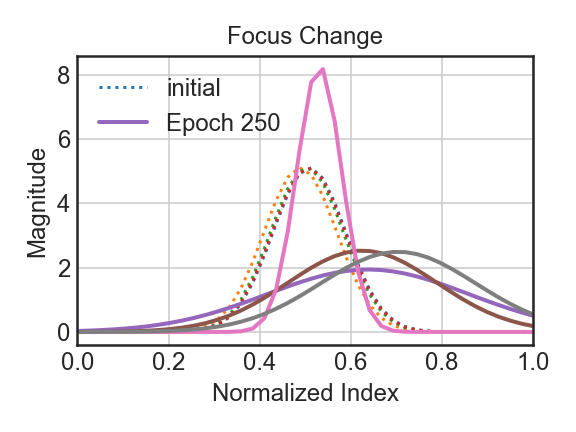}}\\
	\subfloat[]{\includegraphics[width=0.42\linewidth]{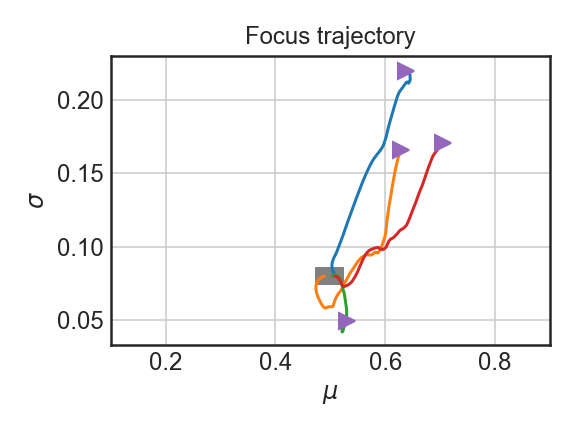}}
	\subfloat[]{\includegraphics[width=0.4\linewidth]{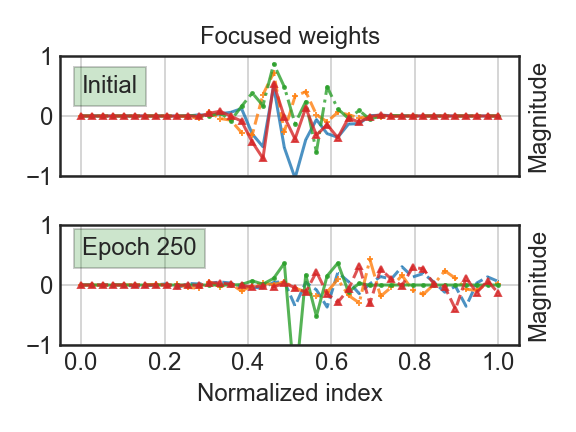}}
	\caption{Locating informative features: a) Input with noise on the sides; foci were initially spread and during the training they shifted to the center. b) Input with noise on the left; foci were initialized in the center and during the training they shifted to the right. c) Trajectory of the foci parameters during training ($\square$: start, $\triangleright$: end). d) Final weights for b).}
	\label{fig:focus_change}	
\end{figure}

A single hidden layer network with four hidden focusing neurons, 2 output neurons, rectified linear activations, and batch normalization was constructed. For the focusing layer, all $\sigma$s were initialized to 0.08. For the left-noised case, $\mu$s were initialized in the center with a small random margin; for the sides-noised data set, $\mu$s were initially spread out over the input space ($[0.2-0.8]$). The focusing neurons' weights $w$ and $\mu$ used a learning rate of $\eta_{\mu,w}=1e-3$, whereas the apertures $\sigma$ used a lower rate of $\eta_\sigma=\eta_{\mu}*0.1$ to train the network with stochastic gradient descent with momentum (0.9) for 250 epochs with a batch size of 128. Figures~\ref{fig:focus_change}a and \ref{fig:focus_change}b show the initial foci and learned foci after 250 epochs for the sides- and left-noised data sets respectively. In the former, since the informative features were in the center, the foci moved in this direction. In the latter, the features were on the right, and the foci accordingly shifted rightwards.

In both cases, most foci were enlarged and migrated towards informative inputs. The apertures were enlarged because only four focusing neurons were used to partition the whole input field. Figure~\ref{fig:focus_change}c shows the changes in foci parameters for the left-noised data set during the training epochs.

In some cases, when the neurons were initialized in pure noise regions (not shown); they could become stuck when the initial aperture was narrow. Therefore, while demonstrating that focusing neurons can seek and focus on informative inputs, this experiment also showed how the same neurons can become jammed in noisy input regions with a narrow aperture. Similar results were observed with larger synthetic input domains and with different settings. Figure~\ref{fig:focus_change}d shows the initial and final focused weights ($\phi_i*w_i$) for each neuron (for the left noise-added data set). The final non-zero weights are apparent on the right and cleaner half of the input space. In the following experiments, the question of whether the demonstrated adaptivity yields an advantage over fully connected neurons was investigated.

\begin{figure}
\centering
	\includegraphics[width=0.5\linewidth]{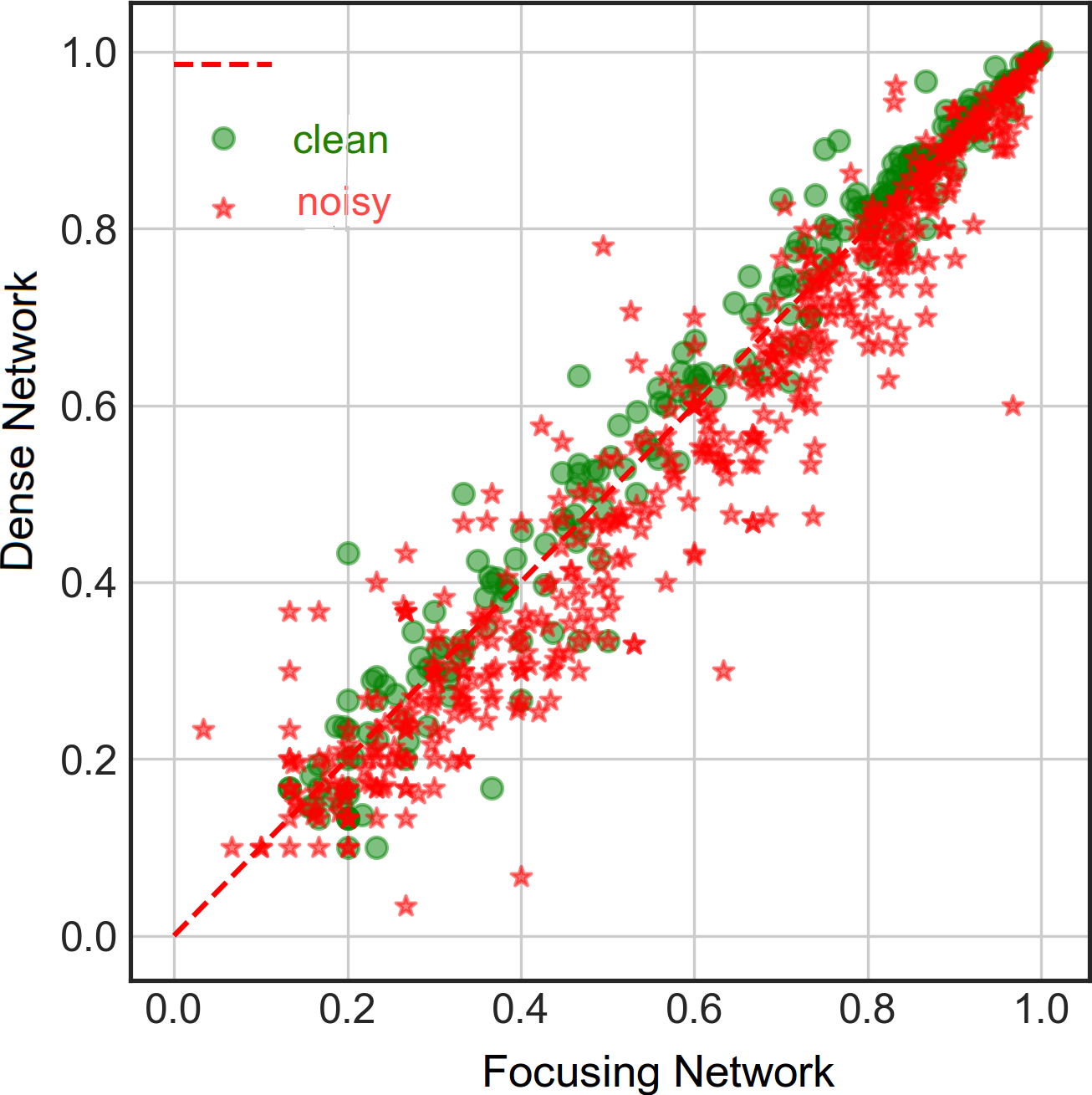}
	\caption{Scatter of test classification accuracies of two-layer focusing
		layer networks (x-axis) versus dense networks (y-axis) obtained in 750 random trials. The parameters for the random data sets and training parameters are given in Table~\ref{tab:params}.}
	\label{fig:scatter}
\end{figure}

To gain more insight, a script was prepared which repeatedly sampled a random synthetic data set, created matching focusing and dense networks, selected random training parameters, and then trained and tested the resulting accuracy of classification in an independent test set of the same distribution. The set of parameters and conditions of the experiment is shown in Table~\ref{tab:params}. Figure~\ref{fig:scatter} shows the test set classification accuracy from 750 runs. Regardless of the data set or training parameters, the focusing networks performed better in the noisy data sets and similar to the dense networks in the clean data sets. A python notebook reproducing and visualizing these results can be found in \citep{gitcode}.

\begin{table}[]
	\centering
	\caption{Synthetic data set, network and training parameter sets}{
	\resizebox{0.5\textwidth}{!}{%
		\begin{tabular}[t]{@{}ll@{}}
			\toprule
			\textbf{\#Input Samples:} & N=\{100,200,2000,10000\} \\ 
			\textbf{Input Length:} & nL=\{4,40,400\} \\
			\textbf{Noisy Dimensions:} & nL*\{0.0,0.5,1.0\} \\
			\textbf{Noise Position:} & \{Sides,Left\} \\
			\textbf{\#Clusters:} & \{1,2,4\} \\
			\textbf{\#Classes:} & \{2,4,8\} \\
			\textbf{\#Hidden Neurons:} & \{4,40,100\} \\
			\textbf{Batch Size:} & N*\{0.1,0.05\} \\
			\textbf{Learning rate:} & \{0.01,0.001\} \\
			\textbf{L2 Penalty:} & \{True,False\} \\ \bottomrule
		\end{tabular}}%
}\label{tab:params}
\end{table}

\subsection{Comparison with Fully Connected Neuron}

\begin{table}[]
	\caption{Test classification performance in spatial 2D data sets. The first part compares the simple (2-hidden) Dense and Focused networks. The second part compares Dense (CNN+Dns) and Focused (CNN+Fcs) layers placed after the flattened convolutional layers. In each case, two tailed t-test results included. N (repeats)=5, p: p-value, highlights indicate \hl{* :p-value<0.05}.}{
	\resizebox{\textwidth}{!}{%
		\begin{tabular}{@{}llccccccccccc@{}}
			\toprule
			& \multicolumn{2}{c}{MNIST} & \multicolumn{2}{c}{CLT} & \multicolumn{2}{c}{CIFAR-10} & \multicolumn{2}{c}{Fashion} & \multicolumn{2}{c}{LFW-Faces} & \multicolumn{2}{c}{Spoken-Digits} \\ \midrule
			
			\multicolumn{6}{l}{\textbf{Simple Two Hidden Layer Networks}}	    &  &  &  &  &  &  & \\ \hline
			
			& Mn$\pm$std & \multicolumn{1}{c|}{Max} & Mn$\pm$std & \multicolumn{1}{c|}{Max} & Mn$\pm$std & \multicolumn{1}{c|}{Max} & Mn$\pm$std & \multicolumn{1}{c|}{Max} & Mn$\pm$std & \multicolumn{1}{c|}{Max} & Mn$\pm$std & Max \\ \midrule

			Dense & 99.02$\pm$1e-3 & \multicolumn{1}{c|}{99.14} & 60.26$\pm$0.01 & \multicolumn{1}{c|}{60.5} & 61.7$\pm$2e-3 & \multicolumn{1}{c|}{62.2} & 90.26$\pm$3e-3 & \multicolumn{1}{c|}{90.98} & 69.85$\pm$4e-3 & \multicolumn{1}{c|}{70.5} & 97.6$\pm$1e-3 & 97.9\\ \midrule
			
			Fixed-s & \textbf{99.25$\pm$6e-5} & \multicolumn{1}{c|}{\textbf{99.28}} & \textbf{73.86$\pm$0.02} & \multicolumn{1}{c|}{\textbf{74.17}} & 63.47$\pm$1e-3 & \multicolumn{1}{c|}{63.87} & 89.83$\pm$7e-4 & \multicolumn{1}{c|}{89.96} & 69.76$\pm$6e-3 & \multicolumn{1}{c|}{70.86} & 97.74$\pm$2e-3 & 97.6 \\ \midrule
			
			Focus-c & 99.11$\pm$2e-3 & \multicolumn{1}{c|}{99.14} & 67.18$\pm$3e-3 & \multicolumn{1}{c|}{67.5} & 64.08$\pm$2e-3 & \multicolumn{1}{c|}{64.34} & 90.9$\pm$2e-4 & \multicolumn{1}{c|}{90.91} & 74.14$\pm$5e-3 & \multicolumn{1}{c|}{75.11} & 98.18$\pm$1e-3 & {98.4} \\ \midrule
			
			Focus-s & \cellcolor[HTML]{B6FF00}99.25$\pm$2e-4 & \multicolumn{1}{c|}{99.27} & {\cellcolor[HTML]{B6FF00}72.1$\pm$0.08} & \multicolumn{1}{c|}{72.22} & \cellcolor[HTML]{B6FF00}\textbf{64.96$\pm$3e-3} & \multicolumn{1}{c|}{\textbf{65.2}} & {\cellcolor[HTML]{B6FF00}\textbf{91.16$\pm$0.07}} & \multicolumn{1}{c|}{\textbf{91.25}} & {\cellcolor[HTML]{B6FF00}\textbf{74.42$\pm$8e-3}} & \multicolumn{1}{c|}{\textbf{75.43}} & {\cellcolor[HTML]{B6FF00}\textbf{98.5$\pm$1e-3}} & \textbf{98.7} \\ \midrule
			
			T-Tests& t & \multicolumn{1}{c|}{p} & t & \multicolumn{1}{c|}{p} & t & \multicolumn{1}{c|}{p} & t & \multicolumn{1}{c|}{p} & t & \multicolumn{1}{c|}{p} & t & p \\ \midrule
			
			
			\begin{tabular}[c]{@{}l@{}}Dense-\\ Focus-s\end{tabular} &  {\cellcolor[HTML]{B6FF00}-6.26}                 & \multicolumn{1}{c|}{\cellcolor[HTML]{B6FF00}2e-4*}                    &   \cellcolor[HTML]{B6FF00}-93.94                & \multicolumn{1}{c|}{\cellcolor[HTML]{B6FF00}2e-13*}                        &\cellcolor[HTML]{B6FF00} -16.1                 & \multicolumn{1}{c|}{\cellcolor[HTML]{B6FF00}2.2e-7*}                       & \cellcolor[HTML]{B6FF00}-5.89                 & \multicolumn{1}{c|}{\cellcolor[HTML]{B6FF00}3e-4*}
			&  \cellcolor[HTML]{B6FF00}-10.1  &
			\multicolumn{1}{c|}{\cellcolor[HTML]{B6FF00}7.8e-6*}&
			\multicolumn{1}{c}{\cellcolor[HTML]{B6FF00}-7.9}  &
			\multicolumn{1}{c}{\cellcolor[HTML]{B6FF00}4.8e-5*}                    \\ \midrule
			\\[0.1cm]
			
			\multicolumn{4}{l}{\textbf{Convolutional Networks}}	 & &   &  &  &  &  &  &  & \\ \hline
			& Mn$\pm$std & \multicolumn{1}{c|}{Max} & Mn$\pm$std & \multicolumn{1}{c|}{Max} & Mn$\pm$std & \multicolumn{1}{c|}{Max} & Mn$\pm$std & \multicolumn{1}{c|}{Max} & Mn$\pm$std & \multicolumn{1}{c|}{Max} & Mn$\pm$std & Max \\ \midrule
			
			CNN+Dns & \textbf{99.59$\pm$2e-4} & \multicolumn{1}{c|}{\textbf{99.63}} & 94.32$\pm$6e-3 & \multicolumn{1}{c|}{94.88} & 77.06$\pm$1e-3 & \multicolumn{1}{c|}{77.69} & \textbf{94.09$\pm$0.01} & \multicolumn{1}{c|}{\textbf{94.28}} & 88.5$\pm$5e-3 & 89.29 & \textbf{98.84$\pm$1e-3} & \textbf{99.0} \\  \midrule
			
			CNN+Fcs & 99.59$\pm$0.02 & \multicolumn{1}{c|}{\textbf{99.63}} & \cellcolor[HTML]{FFE97F}\textbf{96.28$\pm$0.01} & \multicolumn{1}{c|}{\textbf{96.47}} & \cellcolor[HTML]{FFE97F}\textbf{78.71$\pm$0.4} & \multicolumn{1}{c|}{\textbf{79.34}} & 94.06$\pm$0.08 & \multicolumn{1}{c|}{94.19} & {\textbf{89.07$\pm$4e-3}} & \textbf{89.60} & 98.4$\pm$1e-3 &98.6 \\ 
			\midrule

			T-Tests & t & \multicolumn{1}{c|}{p } & t & \multicolumn{1}{c|}{p } & t & \multicolumn{1}{c|}{p }& t& \multicolumn{1}{c|}{p } &t  &\multicolumn{1}{c|}{p} & t & \multicolumn{1}{c}{p }\\ \midrule 
			
			\begin{tabular}[c]{@{}l@{}}Cnn+Dns-\\ Cnn+Fcs\end{tabular} &  +0.52 & 				\multicolumn{1}{c|}{0.61}                      &   \cellcolor[HTML]{FFE97F}-6.4                & 				\multicolumn{1}{c|}{\cellcolor[HTML]{FFE97F}2e-4*}                        &\cellcolor[HTML]{FFE97F} -6.52                 & 				\multicolumn{1}{c|}{\cellcolor[HTML]{FFE97F}1.8e-4*}                       
			& +0.33                 & 				\multicolumn{1}{c|}{0.74}                        
			& -1.58                 & 				\multicolumn{1}{c|}{0.15}                        
			& \multicolumn{1}{c}{\cellcolor[HTML]{FFE97F}+2.4}                 & 	\multicolumn{1}{c}{\cellcolor[HTML]{FFE97F}0.04*}                        
			
			\\ \bottomrule
		\end{tabular}}
	}\label{tab:comp2}
\end{table}

\subsubsection{Spatial 2D Data}\label{sec:real}
To test the focusing model in a real data scenario, the popular gray-scale MNIST character recognition data set \citep{lecun1998a} (MNIST) was the first place to start. More challenging data sets were also used: a cluttered version of MNIST data (CLT), comprised of randomly transformed MNIST samples superimposed on cluttered 60x60 backgrounds \citep{jaderberg2015}; the CIFAR-10 general object classification data set which is composed of 32x32x3 RGB images of ten concrete categories such as car, plane, bird, horse, etc. \citep{cifar10}; and the (FASHION) clothes data set which is arranged similarly to MNIST \citep{xiao2017} to include 10 categories such as t-shirt, pullover, and coat. These almost standard data sets had already been separated into training (60000) and test instances (10000). The tests also included the ``Faces in the Wild'' data set (LFW-Faces) \citep{lfw} as a benchmark for face verification. The LFW-Faces set contains 13233 images of 5749 people; in order to reduce the number of output classes, for the current experiments, individuals with less than 20 images were excluded, resulting in a data set of 3023 (2267 train, 756 test) images of 62 people. Finally, the free spoken-digits data set (Spoken-Digits) \citep{fsdd} was also deployed. This set includes 2500 audio files of digits spoken by 5 people (50 of each digit per speaker) which was randomly split into 1250 training and 1250 test instances. 

A fully connected neural network of two hidden layers was then built according to the following structure: Input - Hidden (800) - BatchNorm - DropOut (0.2) - Hidden (800)- BatchNorm - DropOut (0.25) - Output (10 or 62 for Faces data). The network configuration for this structure can be found in the supplementary figures. For the purposes of comparison, the dense hidden layers were replaced by focused layers, and three different focusing network configurations were compared: 1) A focused layer network (\textbf{Focus-s}) with foci were initially spread out ($\mu\!=\!\left[0.2,0.8\right]$) over the input; 2) Foci initialized at the center (\textbf{Focus-c}), and 3) Foci were initially spread out ($\mu\!=\!\left[0.2,0.8\right]$) however not updated during training epochs (\textbf{Fixed-s}). The Fixed-s $\sigma$ was set at 0.1, whereas for other networks, the $\sigma$s were initialized to 0.025.

The learning rate for each data set was manually tuned to elicit the best performance from the dense network. Then, with the exception of the learning rate, all focusing network hyper-parameters were manually tuned to get the best performance from the focused networks. The learning rates for MNIST, CLT, and CIFAR-10 were $\eta_{g}=0.1$, $\eta_{\mu}=0.01$, and $\eta_{\sigma}=0.01$ for the general (non-focus), focus center and aperture learning rates, respectively. For the Fashion data set, the rates were $\eta_{g}=0.1$, $\eta_{\mu}=0.01$, and $\eta_{\sigma}=0.0005$, respectively. Finally, for the LFW-Faces data set the networks used a single learning rate of $\eta_{g}=0.01$ for all parameters. 

All networks were trained for 200 epochs. The training and test cycle was repeated five times. The averages of the best test accuracies that were calculated using Algorithm \ref{algo:3} over five repeats and their maxima are shown in Table~\ref{tab:comp2}. Other details can be found in the source code provided \citep{gitcode}.

\begin{algorithm}
	\DontPrintSemicolon
	\KwIn{network, dataset, $N_{repeats}$, $N_{epochs}$, Lrrates=$\{\eta_{g}, \eta_{\mu}, \eta_{\sigma}\}$: learning rates for global variables, $\mu$ and $\sigma$ respectively
	Mmrate=$\{m_{g}=0.9\}$: momentum rate.}
	\KwOut{BestTestResults: list of best test accuracies.}
	\Begin{
		BestTestResults = []\;
		\For{$r \leftarrow 0$ \KwTo $N_{repeats}-1$}{
			EpochAccuracyList = []\;
			trainX,trainy,testX,testy = randomsplit(dataset)\;
			\For{$e\leftarrow 0$ \KwTo $N_{epochs}-1$}{
				
				\For{each batch (Xinputs, targets) in (trainX, trainy)}{
					params $\leftarrow $ network.trainableparams\;
					pred $\leftarrow $ network.output(Xinputs)\;
					loss $\leftarrow $ categoricalrossentropy(targets, pred)\;
					
					updates $\leftarrow $ SGD(loss, params, Lrrates, Mmrate)\;
					\emph{//SGD: stochastic gradient descent with momentum}\;
					\If{$type$(network) == focused} {
						updates.append(clip(params.sigma, 0.01, 1.0))\;
						updates.append(clip(params.mu, 0.0, 1.0)) \;
						}
					
					network.update(updates)
					
				}
			
			score $\leftarrow $ accuracyscore(network, testX,testy)\;
			EpochAccuracyList.append(score)
			}
		maxscore $\leftarrow $ max(EpochAccuracyList)\;
		BestTestResults.append(maxscore)\;
			
		}
	}
	\caption{Train, test, and optimization procedure}
	\label{algo:3}
\end{algorithm}
 
\begin{figure}
	\centering
	
	\subfloat[MNIST]{
		\includegraphics[width=0.32\linewidth]{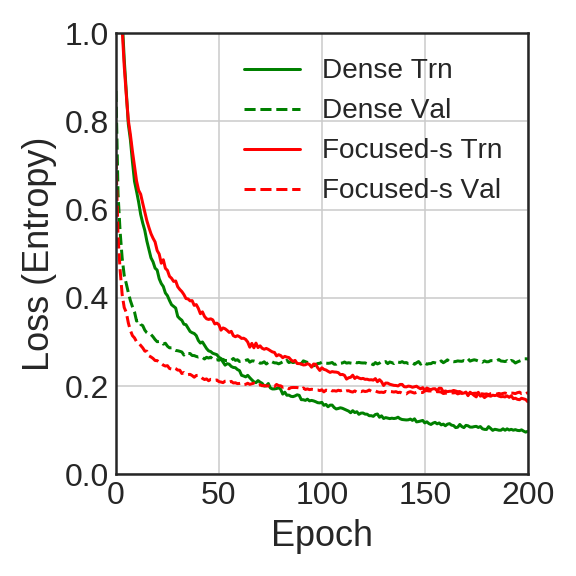}}
	\subfloat[MNIST-CLUTTERED]{
		\includegraphics[width=0.32\linewidth]{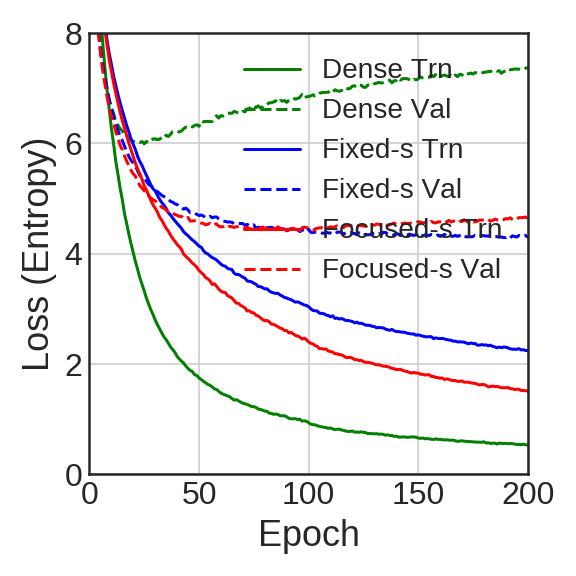}}
	\subfloat[CIFAR-10]{
		\includegraphics[width=0.32\linewidth]{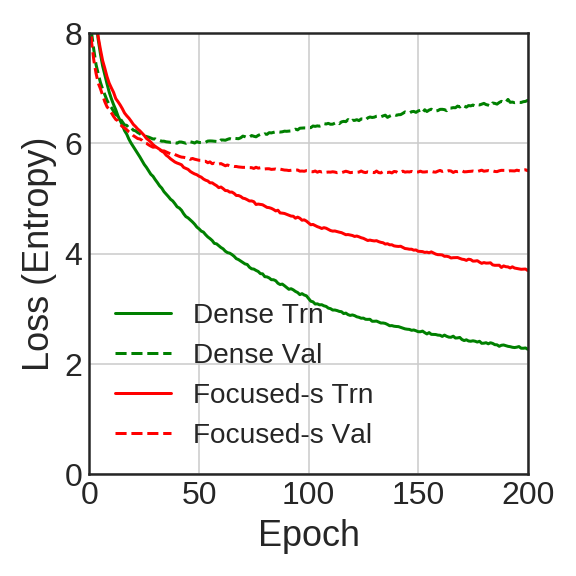}}\\
	\subfloat[FASHION]{
		\includegraphics[width=0.32\linewidth]{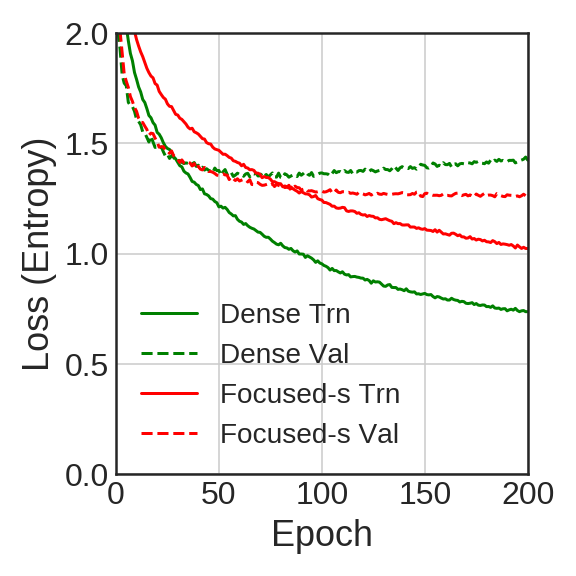}}
	\subfloat[LFW-FACES]{
		\includegraphics[width=0.32\linewidth]{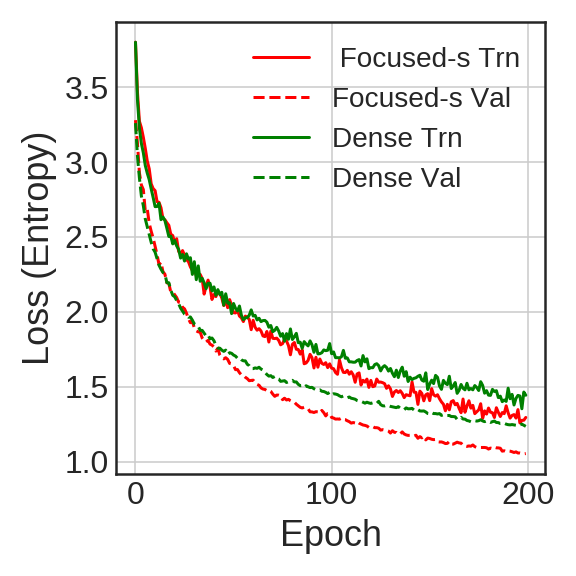}}
	\subfloat[FSDD-Audio]{
		\includegraphics[width=0.32\linewidth]{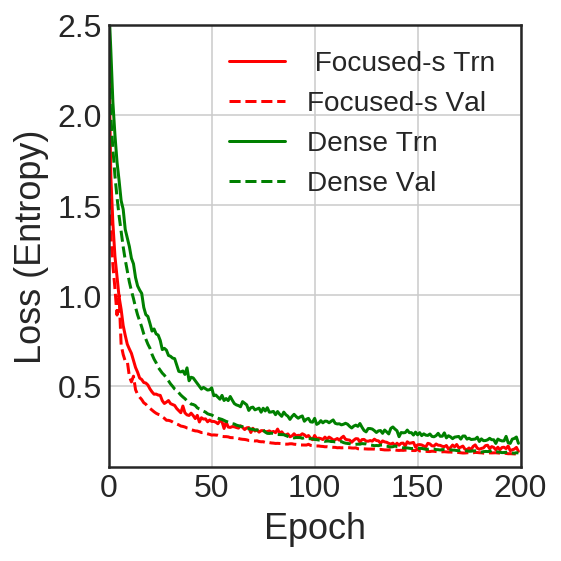}}\\
	\subfloat[REUTERS]{
		\includegraphics[width=0.32\linewidth]{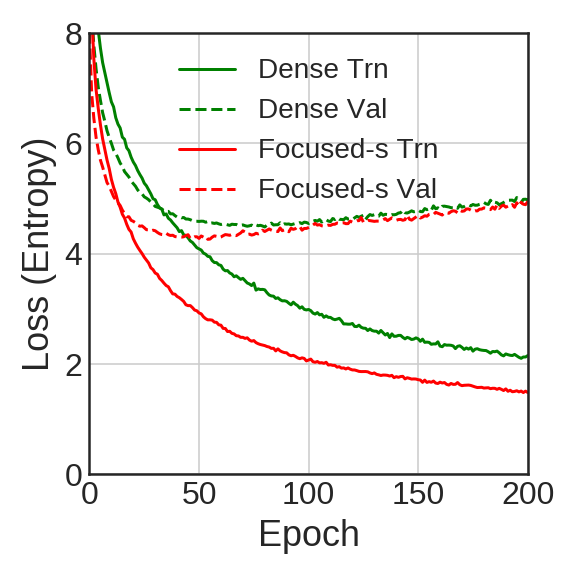}}
	\subfloat[BOSTON]{
		\includegraphics[width=0.32\linewidth]{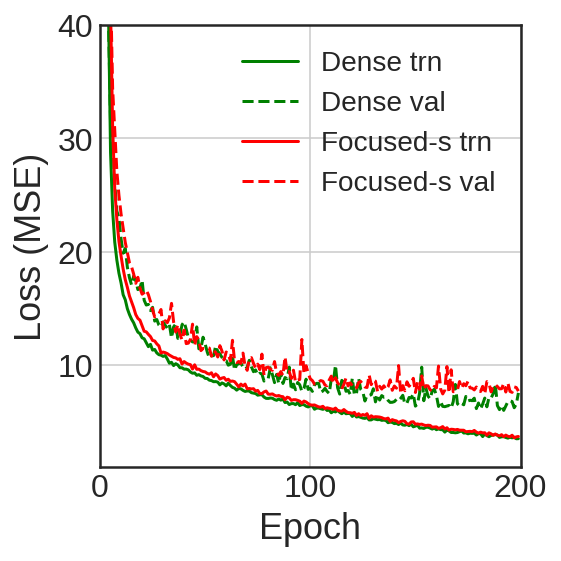}}
	\subfloat[DNA]{
		\includegraphics[width=0.32\linewidth]{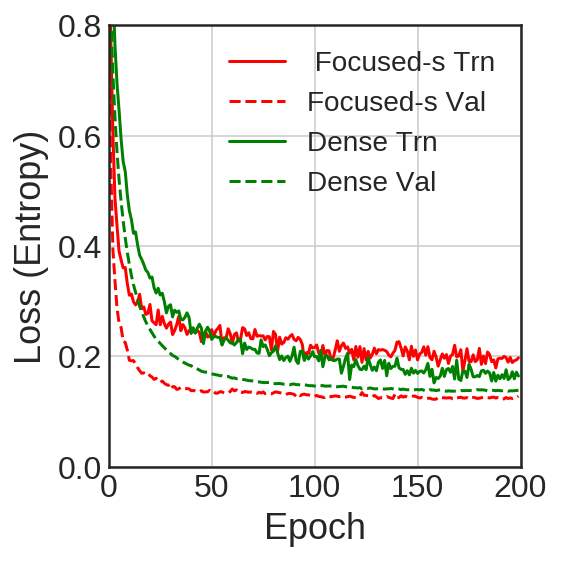}}
	
	\caption{Training (Trn) and Validation (Val) categorical cross entropy (mean squared error for BOSTON) per training epoch of Dense and Focused-s networks for different data sets.}
	\label{fig:losses}
\end{figure}

\emph{MNIST}: For the MNIST data set, all three focused networks performed slightly better than the dense network. Fixed-s was the best performing, with the Focus-s performance almost as effective, and the Focus-c slightly better than the dense network. Figure \ref{fig:losses}a shows the training and test errors that occurred during training iterations (for the top performed dense and Focus-s networks). The dense network reduced the training error rate better; however, the test error rate increased gradually.

\begin{figure}
	\centering
	\subfloat[]{\includegraphics[width=0.5\linewidth]{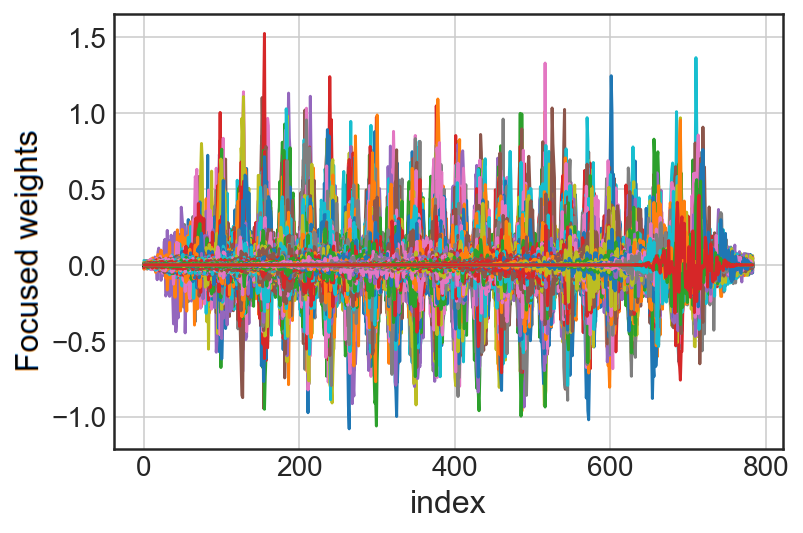}\label{fig:mnist_fi_w_all}}
	\subfloat[]{\includegraphics[width=0.5\linewidth]{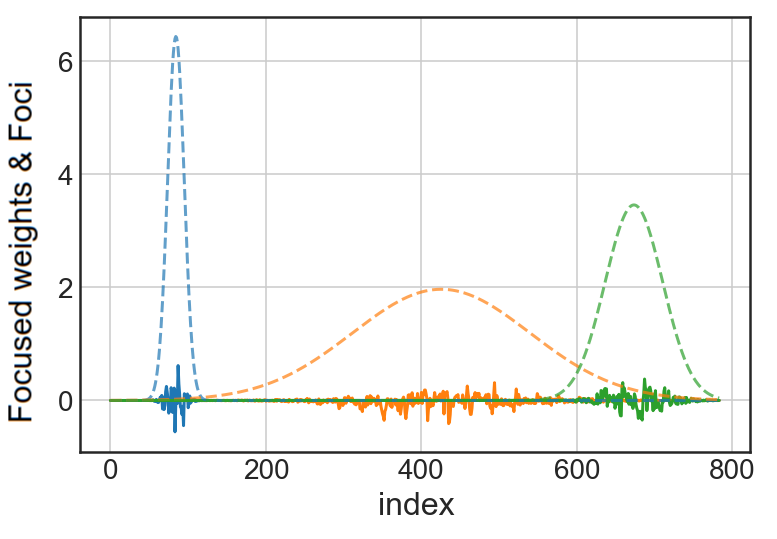}\label{fig:mnist_fi_w}}\\	
	\subfloat[]{\includegraphics[width=0.4\linewidth]{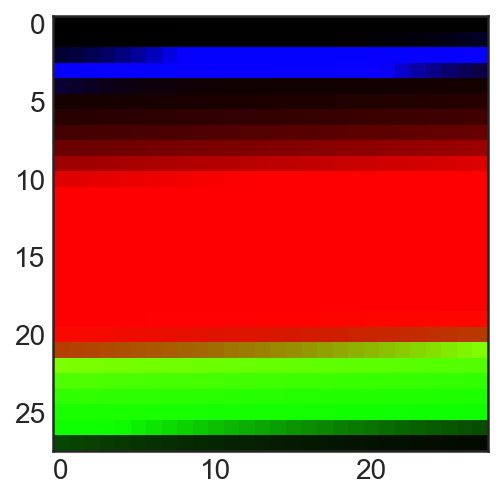}\label{fig:mnist_fi_2d}}
	\subfloat[]{\includegraphics[width=0.4\linewidth]{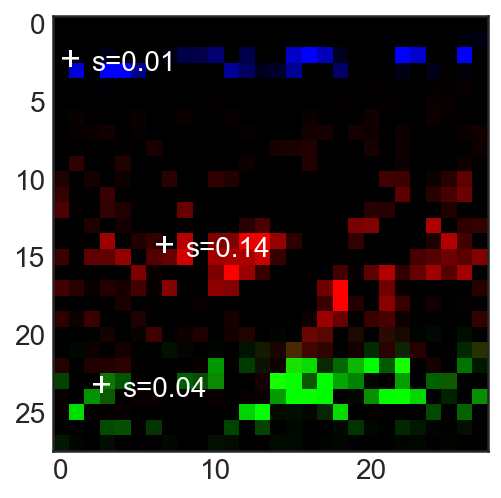}\label{fig:mnist_fi_w_2d}}
	\caption{Locality of the Focus-s network after the MNIST training. a) Focused weights of all neurons in the first layer, b) Foci and focused weight values for three different neurons. c) 2D projection of the focus functions of the same three neurons, d) 2D projection of the focused weights of the same three neurons. + indicates focus center and s denotes $\sigma$ value.}
	\label{fig:mnist_locality}
\end{figure}

To inspect the locality of the connections, Figure~\ref{fig:mnist_fi_w_all} details the focused weights ($w_i*\phi_i$) of all neurons from the first layer of the trained Focused-s network with three example focus functions and focused weights shown in Figure~\ref{fig:mnist_fi_w}. Figures~\ref{fig:mnist_fi_2d} and ~\ref{fig:mnist_fi_w_2d} present 2D projections of the focus functions and the focused weights of the same three neurons. It can be seen that the receptive fields of the first-layer neurons formed overlapping bands in the 2D input domain. 

Figure~\ref{fig:sigma_dist_simple} shows the distribution of the focus apertures ($\sigma$) of the first and second layers of the network after training for 200 epochs. Although all neurons were initialized with a relatively narrow $\sigma$ value (0.025), the average $\sigma$ was close to 0.1 at the end of the training. A video sequence showing the change of the foci during the training can be found in the supplementary materials.

\emph{MNIST-Cluttered (CLT)}: Compared to the original MNIST collection, the images in this data set feature strong background clutter and positional variations inside the 60x60 frame. Table~\ref{tab:comp2} shows that all three configurations of the focused network performed 5-10\% better than the dense network on this set. Surprisingly, the best result was obtained by Fixed-s, which performed slightly better on the task than Focus-s. By checking the validation error rate in Figure~\ref{fig:losses}b, one can see that both the dense network and its focused counterpart overfitted, though this was more prominent in the former case. The Fixed-s network performed slightly better than the trainable focus although it should be noted that the focus parameters ($\mu$=spread, $\sigma$=0.1) were manually tuned for Fixed-s.

\emph{CIFAR-10}: All three configurations of the focused networks performed better than the dense network with the Focus-s network obtaining the highest level of accuracy. Again, the training and validation errors in Figure~\ref{fig:losses}c indicate overfitting in the dense model.

\emph{Fashion}: The highest accuracy was obtained by the Focus-s network. The Focus-c network performed slightly better than the Fixed-s and dense networks. As before, the dense network reduced the training error rate more effectively; but started overfitting earlier than the focused network, as shown by the increased errors in the validation set (Figure~\ref{fig:losses}d).

\emph{LFW-Faces}: In the training of this data set, a data augmentation composed of instance mirror and random xy-translation was used. Again, the Focused-s performed the best, followed by the Focus-c and dense networks. Figure~\ref{fig:losses}e demonstrates that both training and validation errors were reduced further than the Dense network was able to achieve.

\emph{Spoken-Digits}: Due to the limited size of the samples in this data set, 10-fold cross-validations were performed. The audio samples were transformed into Mel frequency cepstral coefficients using the Librosa library. Table~\ref{tab:comp2} shows that the Focused-s network test set performance was the best followed by the Focus-c network. Figure~\ref{fig:losses}e shows lower training and validation errors for Focused-s. 

\emph{T-Tests}: 
To test the significance of the difference in the mean test accuracies two sample t-tests were performed. Table~\ref{tab:comp2} presents the difference between the Focus-s and Dense network mean accuracies. Across all data sets, there were statistically significant differences in favor of the Focused-s network (highlighted green). The improvements recorded across the more challenging data sets were particularly remarkable: CLT $\approx12\%$, CIFAR-10 $\approx3.4\%$, LFW-Faces $\approx4.5\%$.

\subsection{Number of Hidden Layer Neurons}
\begin{figure}
\centering
\subfloat[MNIST]{\includegraphics[width=0.75\linewidth]{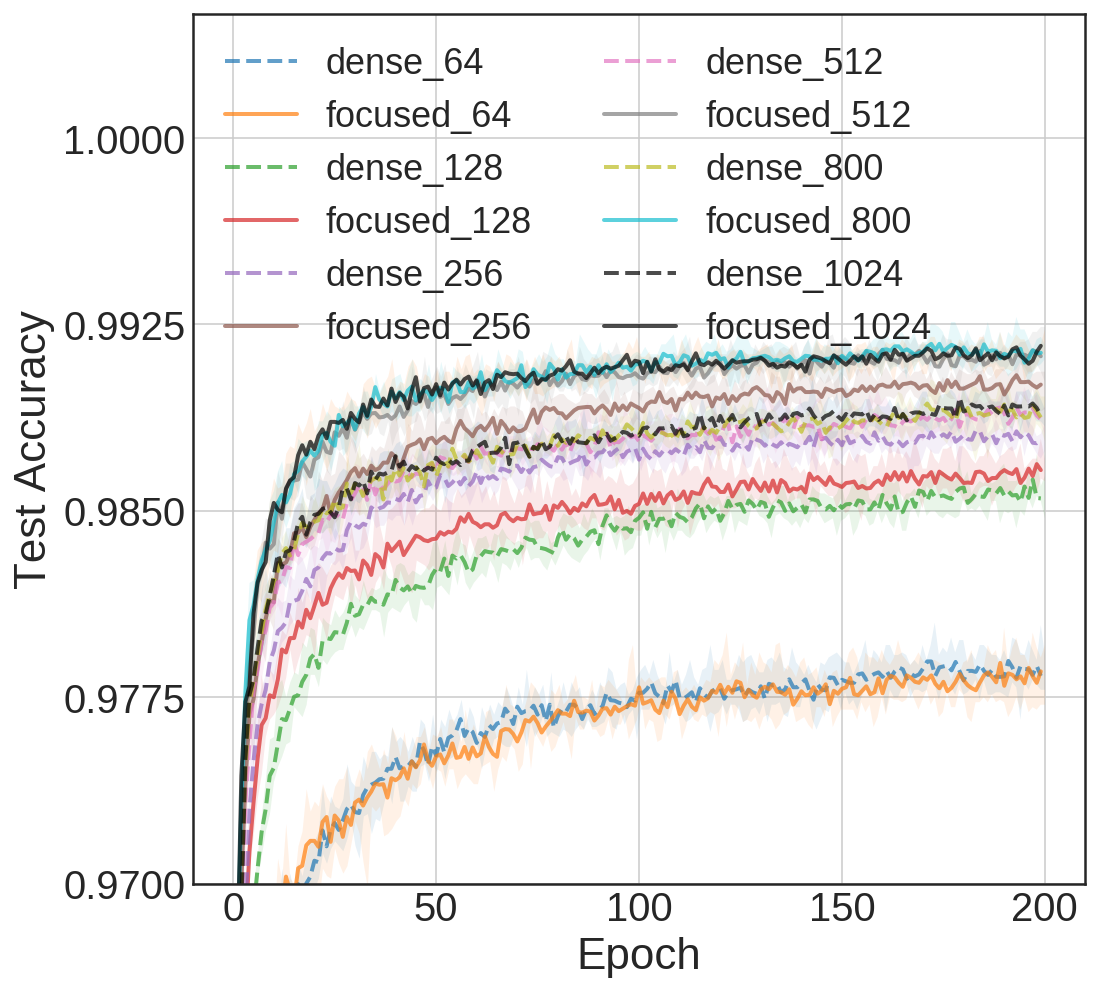}}\\
\subfloat[CIFAR-10]{\includegraphics[width=0.75\linewidth]{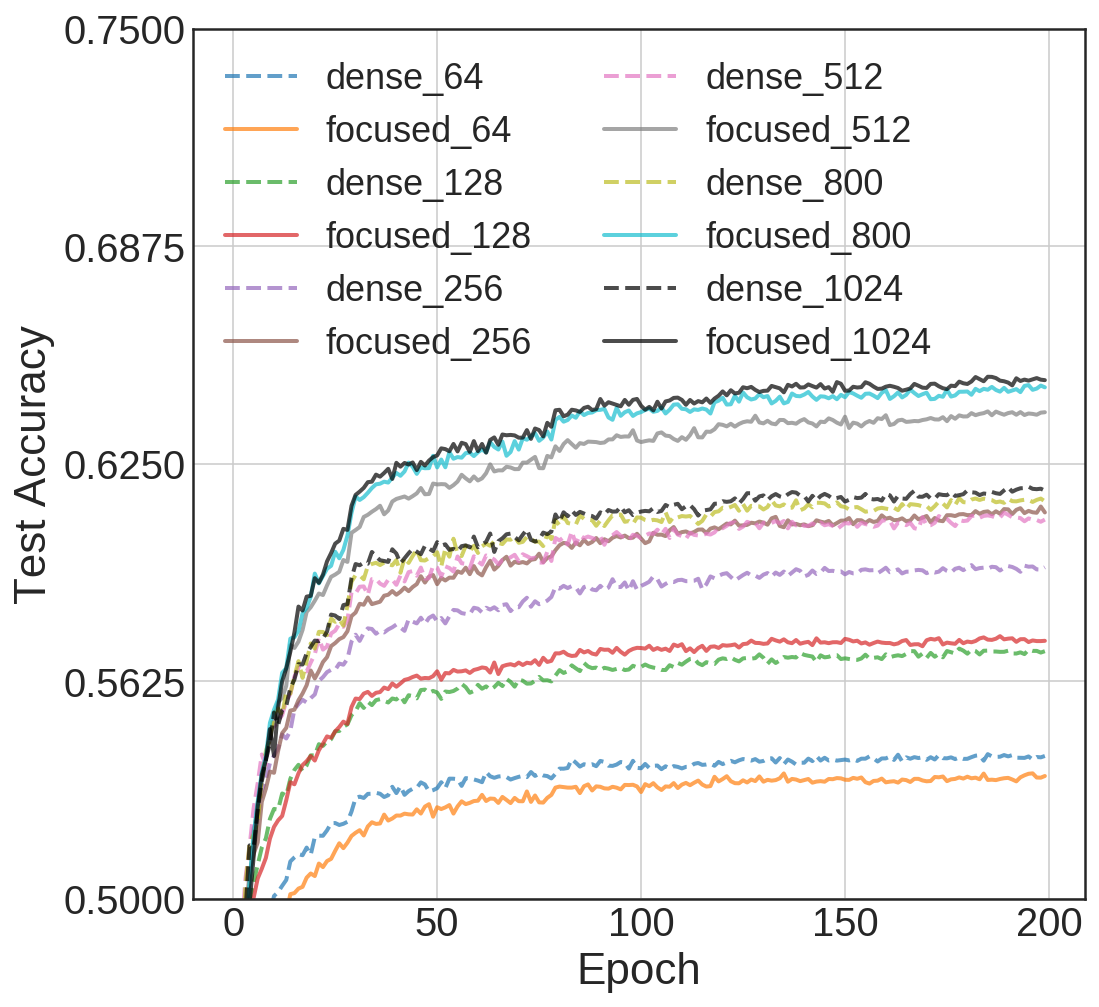}}
\caption{Comparison of the mean test accuracies of the Focus-s and Dense networks with different number of hidden neurons (64, 128, 256, 512, 1024).}
\label{fig:nhidden}
\end{figure}

The next experiment sought to compare the networks when the number of neurons in the hidden layers was changed. To this end both networks (Focus-s and Dense) were reconstructed with different numbers of neurons in the hidden layers $\{64, 128, 256, 512, 800, 1024\}$ when classifying the MNIST and CIFAR-10 sets. For both networks, the train-test cycle was repeated 5 times for each size. The average test accuracies shown in Figure \ref{fig:nhidden} revealed that, except for the lowest size ($64$), the Focus-s networks outperformed their dense counterparts, which used the same number of neurons. Moreover, the Focus-s networks with $256$ or more neurons in the hidden layers outperformed any Dense network tested on the data sets.

\subsection{Sparsity}
As mentioned earlier, the focus function is only necessary during training. At run-time, the product of the focus coefficients and weights can be used instead. Moreover, eliminating out-of-focus weights in a trained network would produce a simplified and sparser network. For the sake of brevity, let us write $\phi_i$ for the focus coefficient $i$. If out-of-focus weights are removed by thresholding $\phi_i$ the trained network can be pruned. Removing out-of-focus weights as described above differs from dropping smaller weights in a network as it was done in \citep{han2015deep}.

\begin{figure}
	\centering
	\subfloat[]{\includegraphics[width=0.5\textwidth]{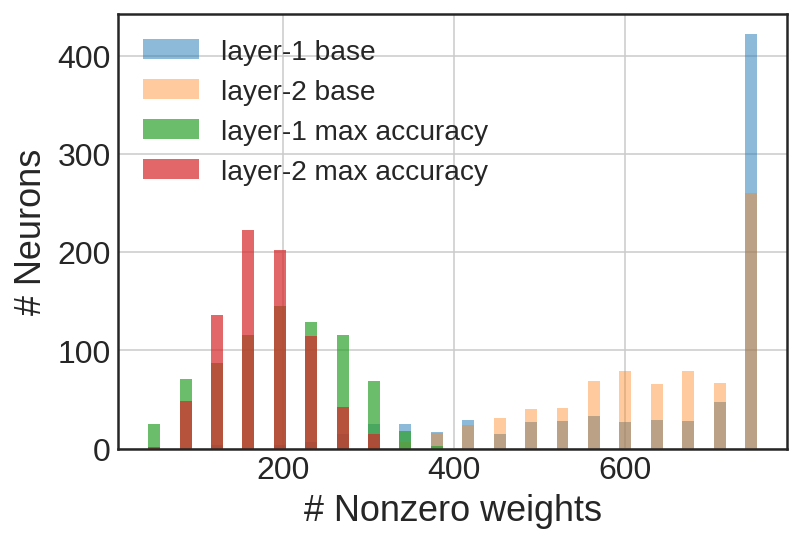}\label{fig:sparsityoflayersmnist10}}
	\subfloat[]{\includegraphics[width=0.5\textwidth]{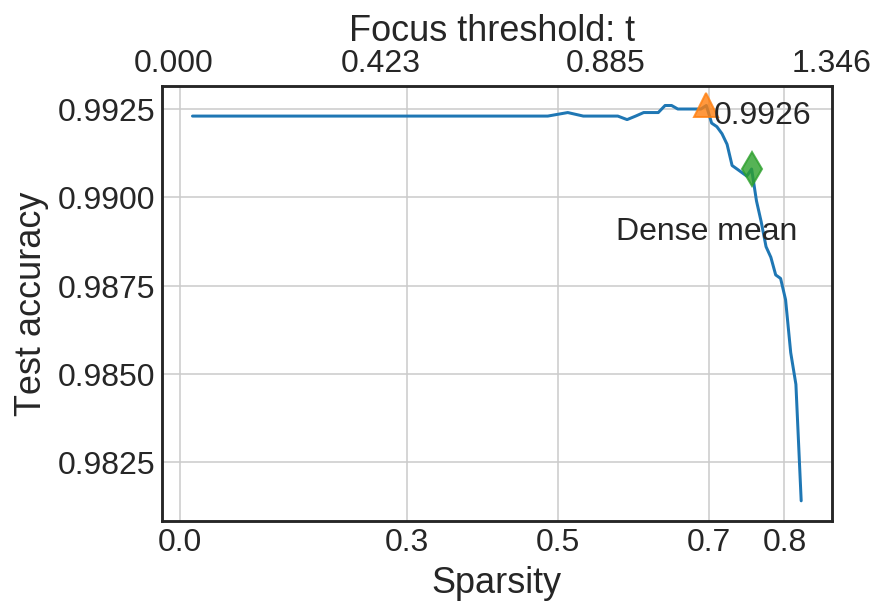}\label{fig:sparsityvsaccuracy}}
	\caption{Focus-s network sparsity after MNIST training. a) Distributions of non-zero connection counts in layer1/layer2 neurons by removing out-of-focus weights with (base: $\phi_i[\phi_i<1e-7]=0$) and (max acc: $\phi_i[\phi_i<1.0]=0$) b) Test set accuracy for increasing sparsity obtained with increasing out-of-focus threshold $t$ in range $[0,1.5]$. The unpruned focused network (base) performance was 99.24 in the MNIST test set.}
	\label{fig:mnist_sparsity}
\end{figure}

This was demonstrated on a focused network (Focused-s) trained on the MNIST data set. Figure~\ref{fig:sparsityoflayersmnist10} shows the distribution of the number of non-zero connections per neuron in the first and second hidden layers achieved by the setting $\phi_i[\phi_i\text{<1e-7}]=0$ and $\phi_i[\phi_i\text{<1.0}]=0$. With the first pruning, the test accuracy was unaffected. In the latter case, the accuracy was increased, although on average, most neurons were connected to a quarter of the inputs. 

In order to establish the tolerance of the network to pruning, accuracy was examined against increased pruning or sparsity. Sparsity was calculated as the ratio of the number of zero weights to the number of total weights. Figure~\ref{fig:sparsityvsaccuracy} plots the test set classification accuracy for increasing sparsity obtained by pruning with an increasing threshold value $t$ ($\phi_i[\phi_i<t]=0$). The accuracy of the pruned network was still above the dense network even when more than $70\%$ of the connections were dropped. Notably, it was also possible to improve accuracy to a level slightly above that of the base level. Blundel et al.\ \citep{blundel_2015} showed that it is possible to remove 98\% of 2.4M weights in a two hidden layer network while maintaining 1.39\% test error on the MNIST set. In the current case, 0.74\% test error was achieved while removing $\approx$70\% of 1.2M weights.

\begin{figure}
	\subfloat[]{\includegraphics[width=0.5\linewidth]{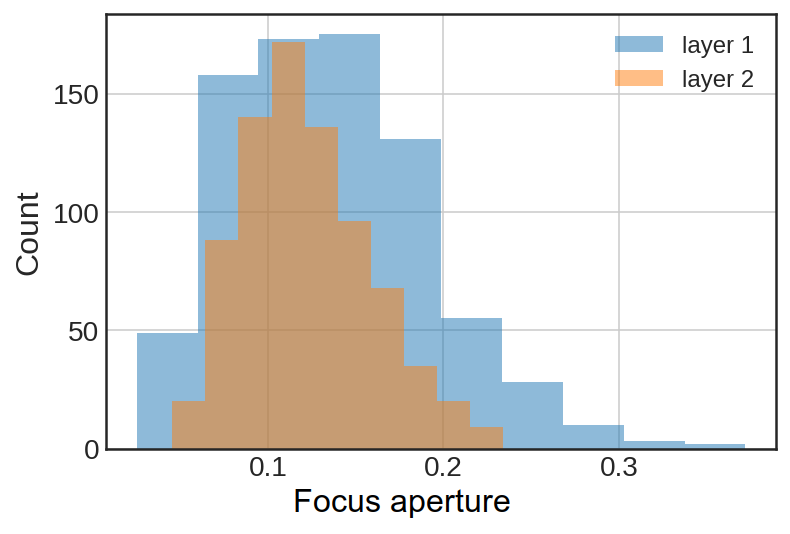}\label{fig:sigma_dist_simple}}
	\subfloat[]{\includegraphics[width=0.5\linewidth]{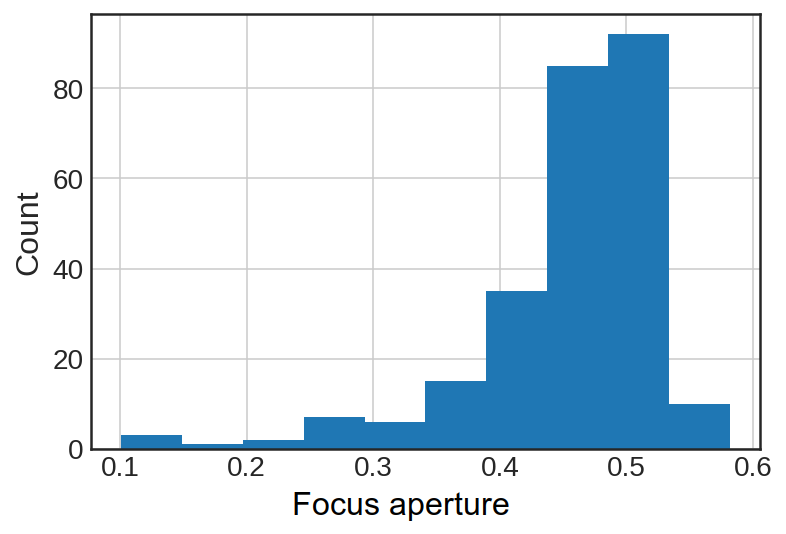}\label{fig:sigma_dist_cnn}}\\
	\caption{Histogram of $\sigma$ values after the training a) for the focused layers in Focus-s and b) in the focused classification layer of the CNN+Fcs network.}
\end{figure}

\subsection{Use as a classification layer}
\subsubsection{Convolutional Networks}
Currently, the most popular use for dense layers is in the classification of deep representations computed by convolutional feature extractors. Here, the role of the dense neuron is different from the ones facing the input in simple 2-hidden layer networks. Therefore, there are two main questions. First, how would a convolutional network perform in the experimental setup with the same data? Second, how would the focusing neuron layers perform in classifying convolutional features? To address these questions, a minimal convolutional network (CNN+Dns: Conv (32) - Conv (32)- Pool (2,2)- Dropout (0.5)- Dense (256) - Dropout (0.5) - Dense (10)) was set up. The network had one more layer than the previously compared networks, however had less parameters (structure provided in supplementary figures). In the MNIST, CIFAR-10, Fashion, CLT tests the learning rates were set as $\{\eta_{g}=0.1, \eta_{\mu}=0.01, \eta_{\sigma}=0.01\}$ respectively for global variables, $\mu$, and $\sigma$. The LFW-Faces and Spoken-Digits tests used single learning rates for all variables: $\eta_{g}=0.01$ and $\eta_{g}=0.05$, respectively.

As before, the training and test cycles were repeated five times. Averages of the maximum test accuracies were calculated as shown in Algorithm \ref{algo:3}. The CNN+Dns row in Table~\ref{tab:comp2} summarizes the results. As expected, the CNNs were superior to both the simple focused and dense networks previously tested. The gain was most prominent in the MNIST cluttered set and LFW-Faces, thanks to the translation invariance provided by the convolutional layers and pooling.

In the counter network a focusing layer replaced the dense layer immediately following the convolutional layers (CNN+Fcs: Conv (32) - Conv (32)- Pool (2,2)- Dropout (0.5)- Focus (256) - Dropout (0.5) - Dense (10)). As the Cnn+Fcs row of Table~\ref{tab:comp2} indicates, the results were almost equal in the MNIST set, in favor of the CNN+Dns in Fashion and Spoken-Digits data sets, and in favor of the focused the CNN+Fcs in the MNIST-cluttered, CIFAR-10, and LFW-Faces. The distribution of apertures in the focused layer (for CIFAR-10) is shown in Figure~\ref{fig:sigma_dist_cnn}. Due to the classification role, the focus apertures opened wider than the ones in the simple network case.

\emph{T-Tests}: T-tests part of Table~\ref{tab:comp2} compares the mean accuracy scores achieved by CNN+Dns with those of CNN+Fcs. A significant difference (highlighted orange) was observed in the MNIST-cluttered and CIFAR-10 data sets in favor of CNN+Fcs and Spoken Digits in favor of the dense counterpart.

\subsubsection{Aperture regularization} 
\begin{figure}
	\centering
	\subfloat[MNIST-Focus-s]{\includegraphics[width=0.4\textwidth]{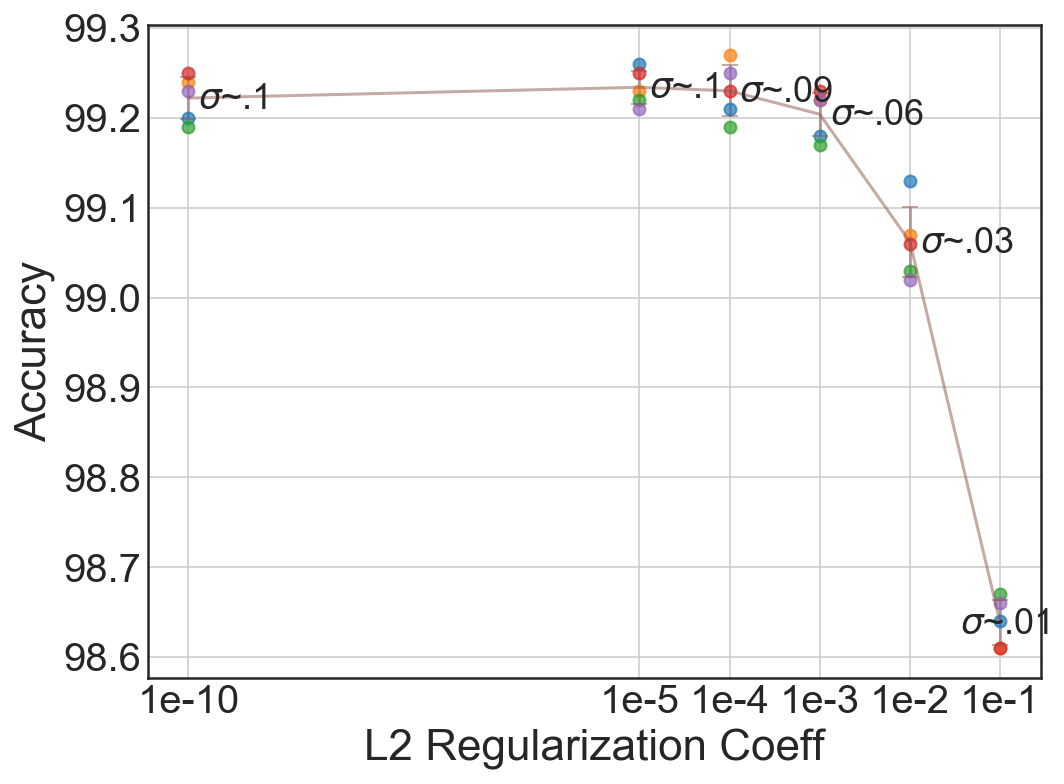}\label{fig:reg_mnist_simple}}
	\subfloat[MNIST-CNN+Fcs]{\includegraphics[width=0.4\textwidth]{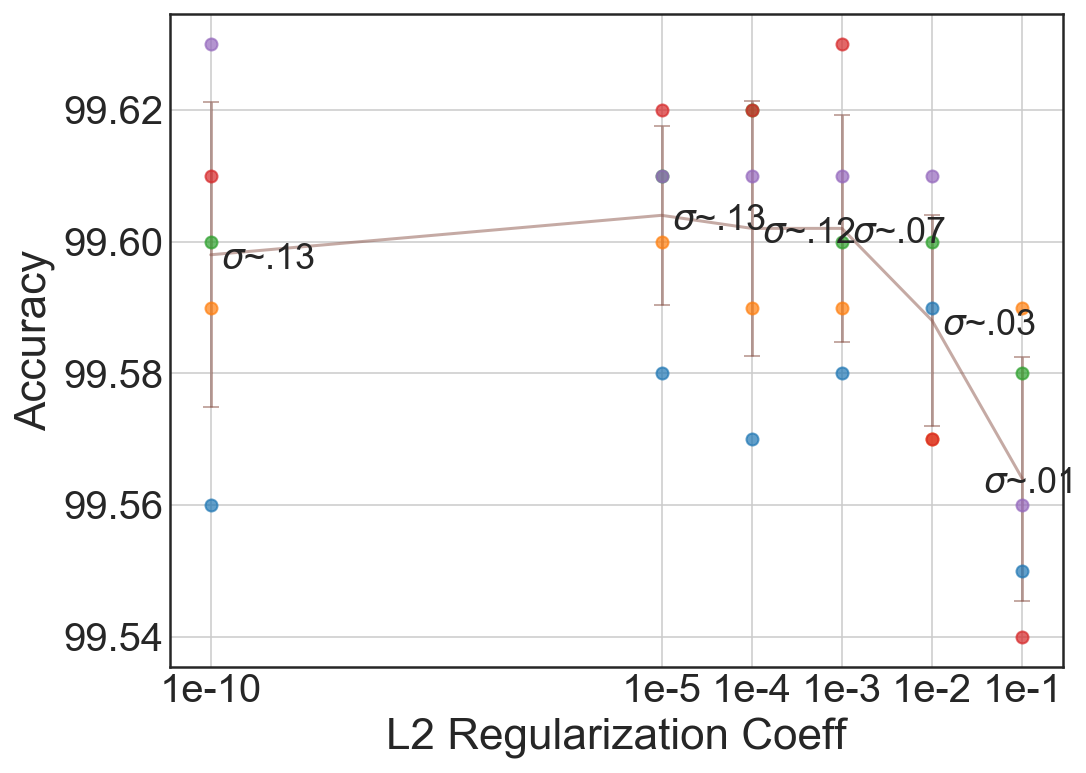}\label{fig:reg_mnist_cnn}}\\
	
	\subfloat[CLT-Focus-s]{\includegraphics[width=0.4\textwidth]{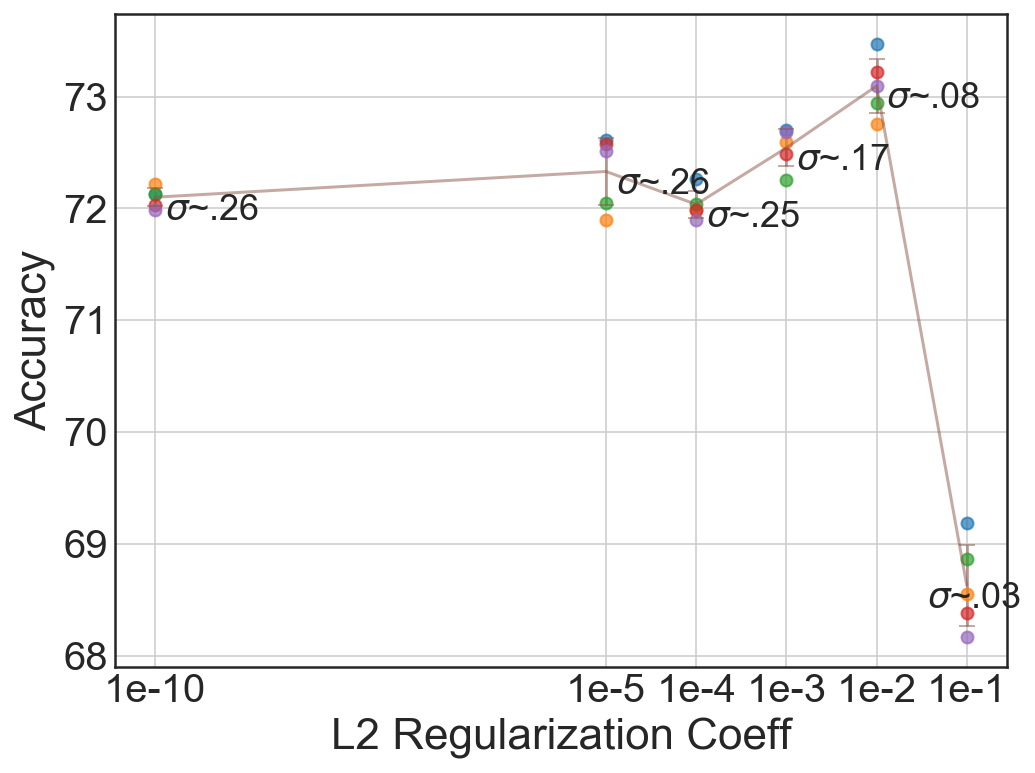}\label{fig:reg_clut_simple}}
	\subfloat[CLT-CNN+Fcs]{\includegraphics[width=0.4\textwidth]{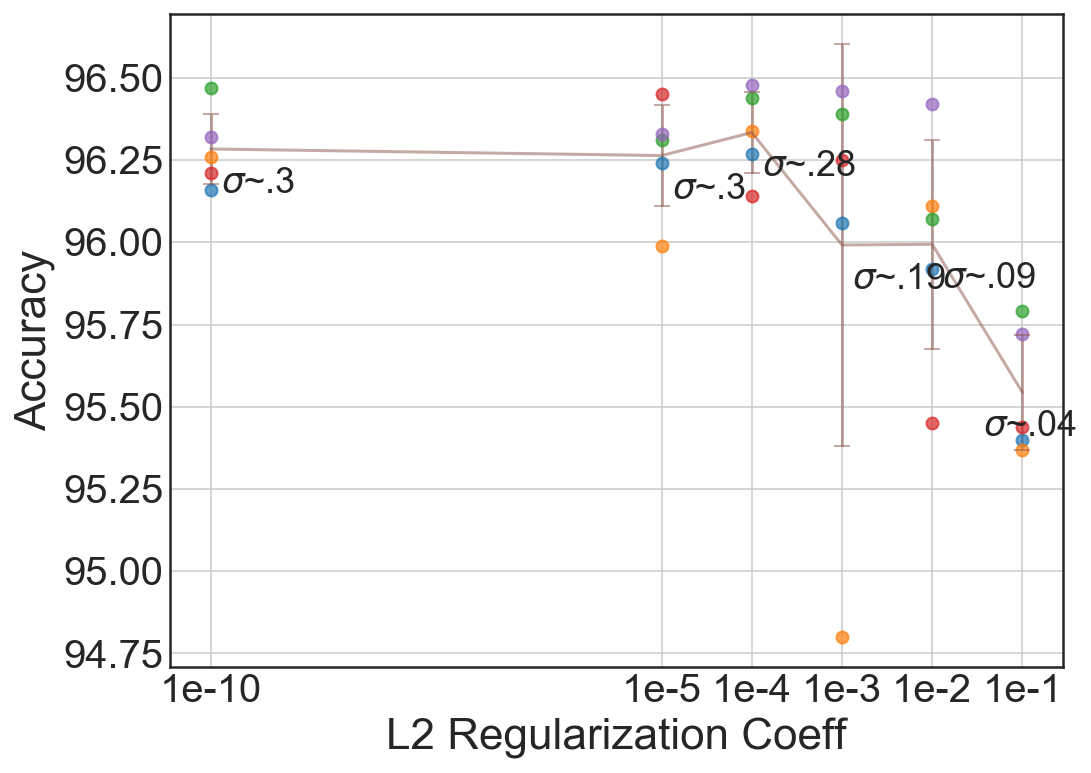}\label{fig:reg_clut_cnn}}\\
	
	\subfloat[CIFAR-Focus-s]{\includegraphics[width=0.4\textwidth]{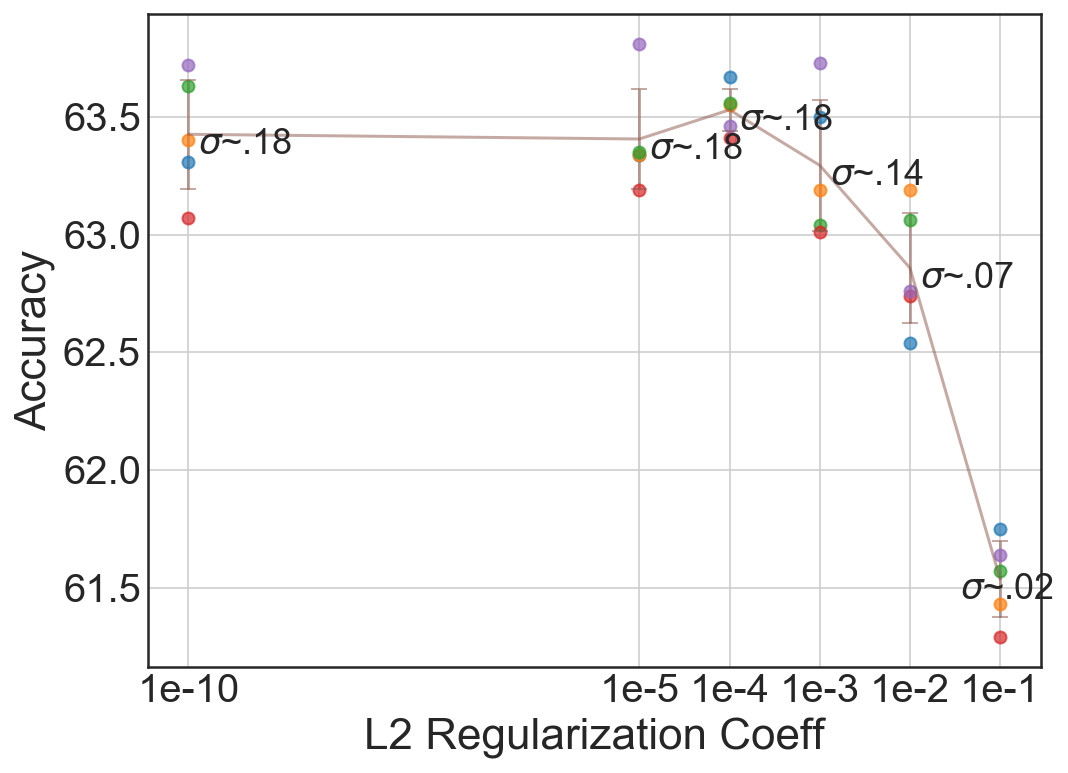}\label{fig:reg_cifar_simple}}
	\subfloat[CIFAR-CNN+Fcs]{\includegraphics[width=0.4\textwidth]{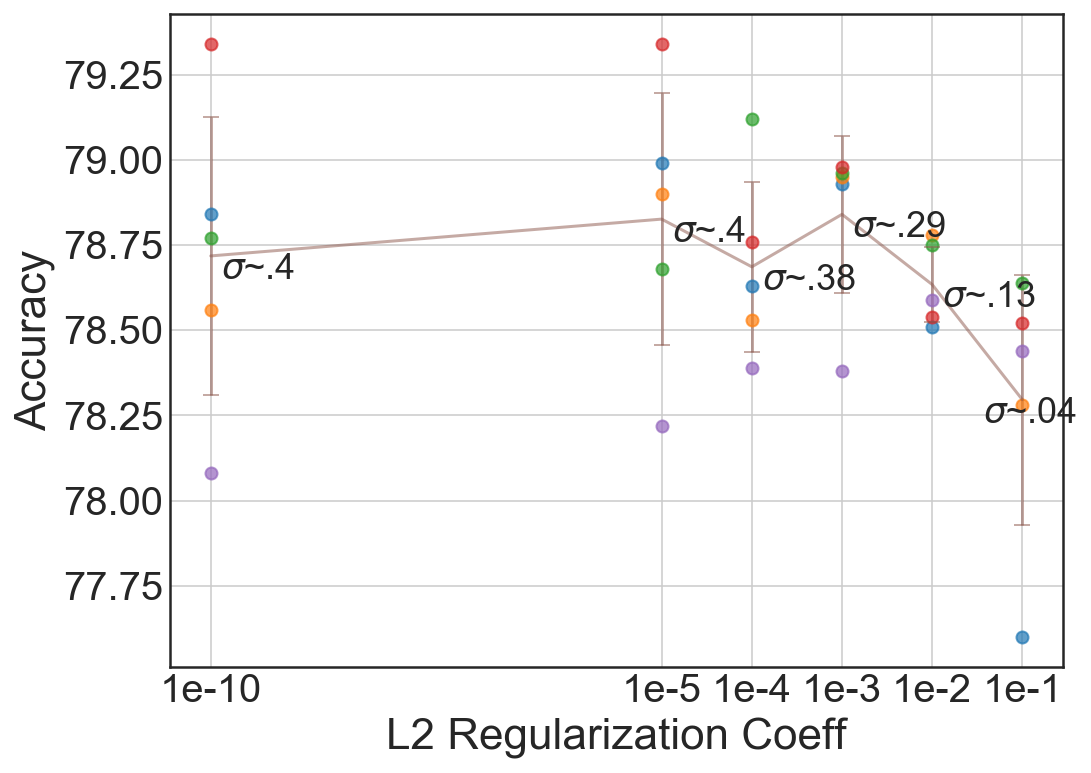}\label{fig:reg_cifar_cnn}}\\
	
	\subfloat[FASHION-Focus-s]{\includegraphics[width=0.4\textwidth]{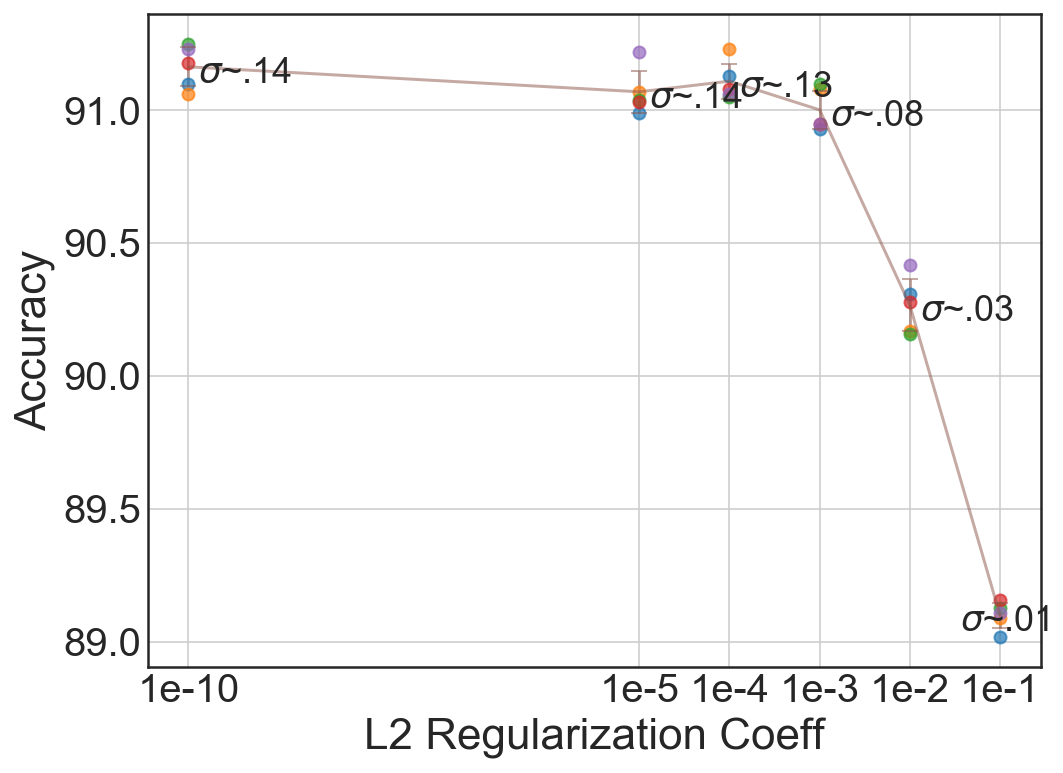}\label{fig:reg_fashion_simple}}
	\subfloat[FASHION-CNN+Fcs]{\includegraphics[width=0.4\textwidth]{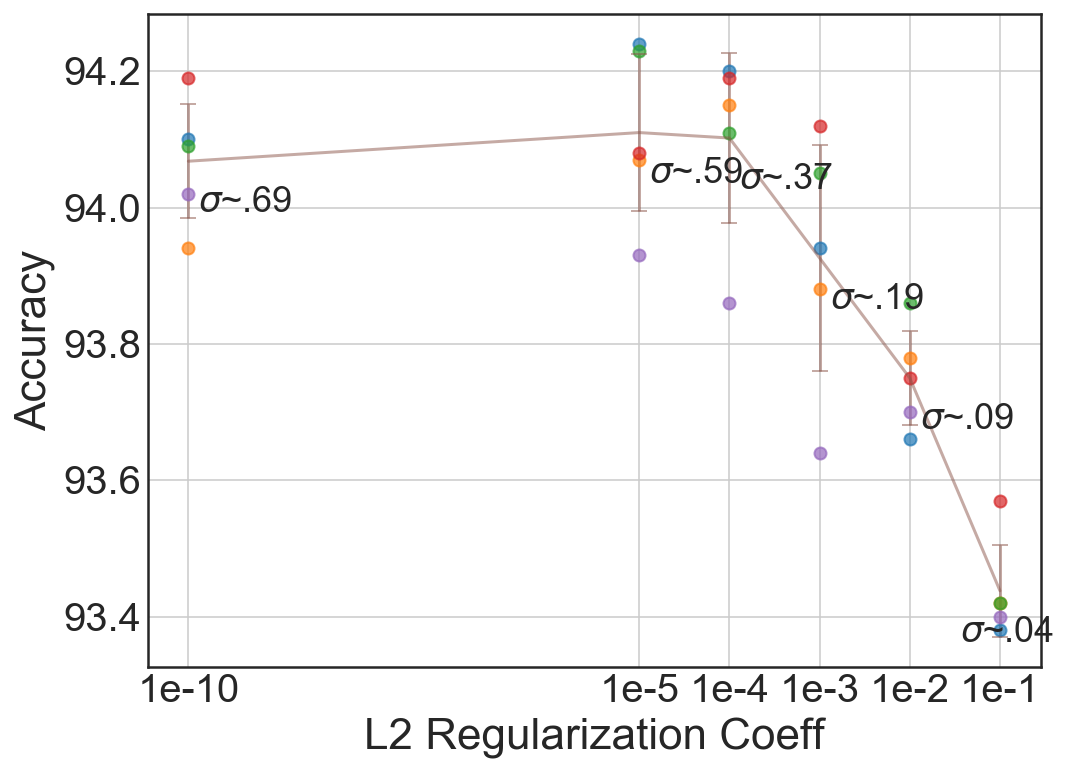}\label{fig:reg_fashion_cnn}}	
	\caption{Regularization of focus aperture $\sigma$ vs test accuracy in different datasets.}
	\label{fig:sigma_reg}
\end{figure}

It is possible to encourage focusing neurons to learn narrow apertures. The effect of L2 regularization of aperture $\sigma$ on the test accuracies on four of the data sets was tested. Figure~\ref{fig:sigma_reg} shows that the same or better results can be obtained by applying regularization coefficients such as 1e-5 or 1e-3 on the aperture parameter $\sigma$, which may promote even sparser networks. 

\subsubsection{Transfer Learning}
Recently, using the pre-trained network weights of successful network architectures (e.g., VGG from Visual Geometry Group \citep{simonyan2014deep} or Resnet \citep{resnet}) as a starting point to learn new data sets has become a feasible approach when training data and/or computational resources are limited, as was the case in this research. Hence, in this study, a focusing layer (40 neurons) was compared against a dense layer (40 neurons) to replace the top classifier layer (CL) of the pre-trained VGG-16 network (VGG16-CL-Output) in the CIFAR-10 and Cats\&Dogs (from Kaggle) data sets. The Cats\&Dogs set includes 19000 training and 6000 test instances of two categories. The images were resized into 125x125 frames, and the training data was augmented with random horizontal flip, random shear and zoom operations. A global average pooling layer was employed to summarize deep convolutional features, which was then fed to the focused or dense layer, followed by an output softmax layer. All the weights, including VGG-16's, were trained using the stochastic gradient descent.

For CIFAR-10, the learning rates of all parameters (including VGG-16's) were set at 1e-3, with the exception of the rate for $\sigma$ which was set at 1e-4. The network was trained in 32 batches for 250 epochs. The focusing layer was initialized with focus centers initially spread out and $\sigma=0.025$. For Cats\&Dogs, the learning rate for $\sigma$ was set at 1e-2; and the network was trained using 64 batches for 50 epochs.

Table~\ref{tab:transfer} demonstrates that the results for both data sets were similar. With p-values 0.38 and 0.47 for CIFAR-10 and Cats\&Dogs respectively, the results could not confirm the statistical significance of the differences.

\begin{table}[]
\centering
	\caption{Comparison of Focused (Vgg+Fcs) vs. Dense layer classification (Vgg+Dns) in transfer learning in CIFAR-10. Mn: mean, std: standard deviation, p: p-value (*significance).}{
	\resizebox{0.75\textwidth}{!}{%
		\begin{tabular}{@{}ccccccc@{}}
			\toprule
			
			\multicolumn{1}{l}{} & \multicolumn{3}{c}{CIFAR-10}  & \multicolumn{3}{c}{Cats\&Dogs} \\		\midrule
			
			& Mn$\pm$std & Max & t(p) & Mn$\pm$std & Max &t(p) \\ \midrule
			
			Vgg+Dns & 91.77$\pm$.08 & \textbf{91.92} &  \multirow{+2}{*}{-0.28 (0.78)} & \textbf{97.6$\pm$.02} & 97.85 &  \multirow{+2}{*}{0.74 1(0.47)}    \\
			
			Vgg+Fcs & \textbf{91.8$\pm$9e-4} & 91.9 &   & 97.46$\pm$3e-3& \textbf{97.88} &    \\		
			\bottomrule
		\end{tabular}}
	}\label{tab:transfer}
\end{table}

\begin{figure}
	\centering
	\subfloat[MNIST]{\includegraphics[width=0.65\textwidth]{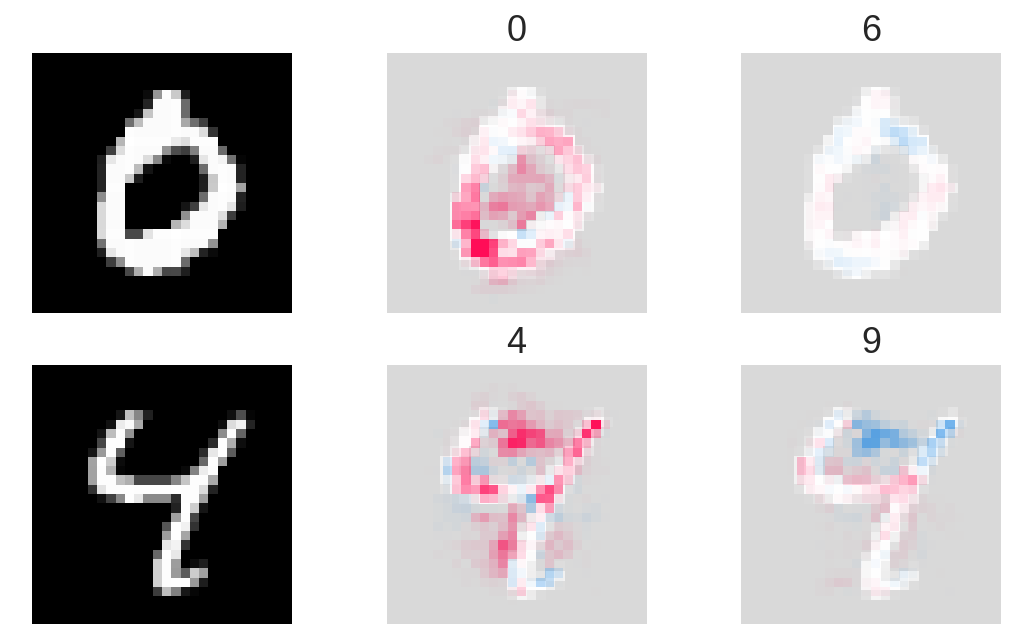}\label{fig:shap_mnist}}\\
	\subfloat[CIFAR10]{\includegraphics[width=0.65\textwidth]{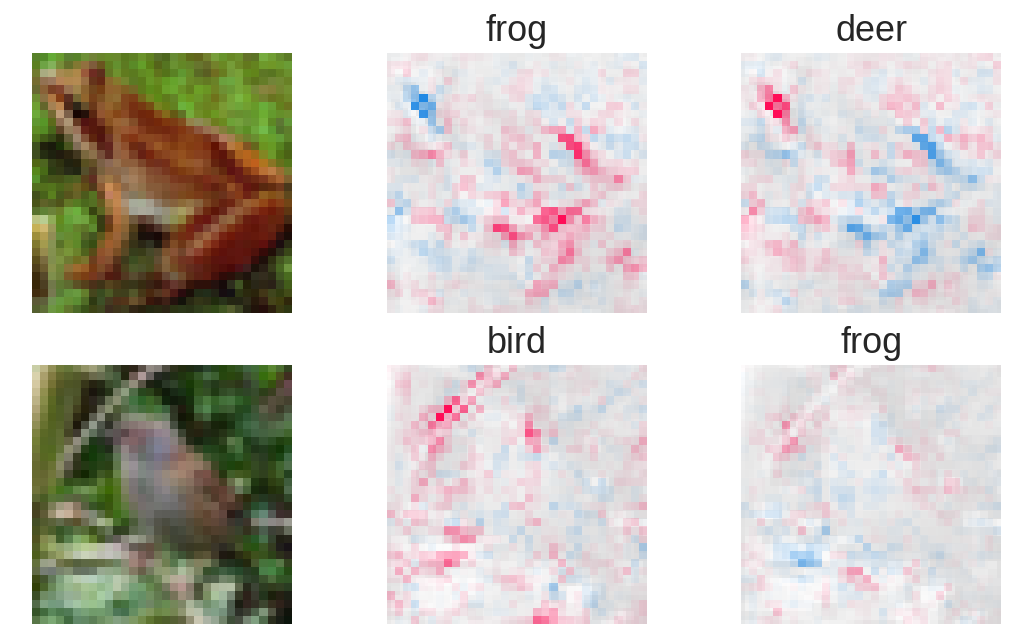}\label{fig:shap_cifar}}\\
	\subfloat[CIFAR10]{\includegraphics[width=0.65\textwidth]{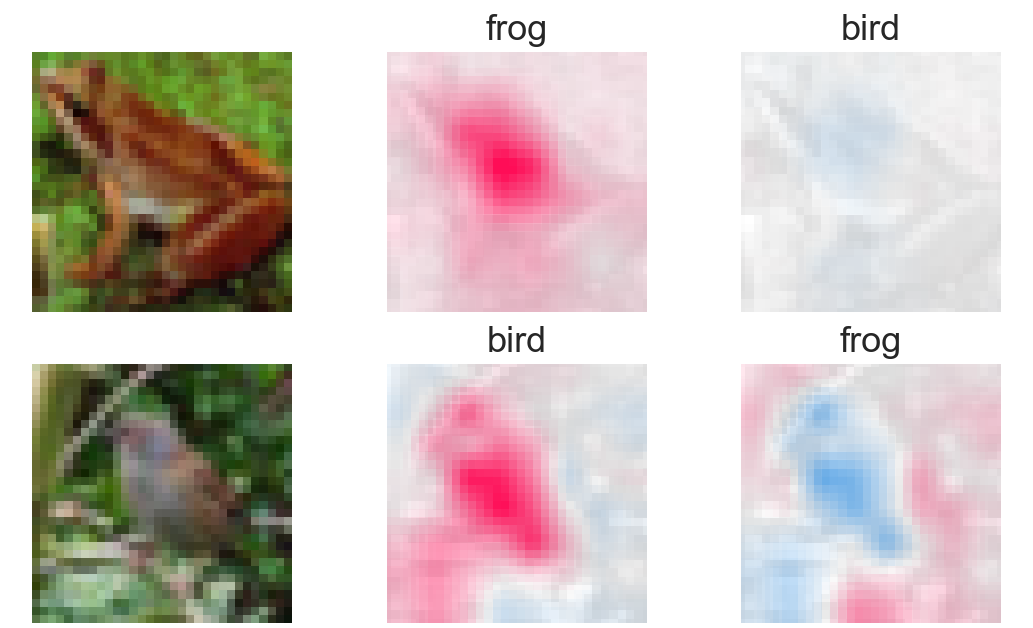}\label{fig:shap_cifar_transfer}}\\
	\subfloat{\includegraphics[width=0.65\textwidth]{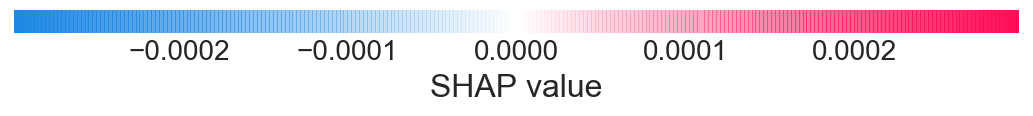}\label{fig:shap_cropped}}	
	\caption{Shap value images for different networks and datasets. a) Focus-s (simple 2-hidden layer) at MNIST classification). b) CNN+Fcs at CIFAR-10 classification, c) VGG+Fcs network at CIFAR-10 classification.}
	\label{fig:shap_values}
\end{figure}

\subsubsection{Where do neurons focus?}
Focusing neurons are static, meaning they do not change their focus per input, and foci are learned from the training data over many iterations. Nevertheless, it remains legitimate to inquire which parts or regions of a particular input contributed most to the network decision. To this end, Shapley additive explanations (SHAP) \citep{lundberg_2017}, which are generalizations of decision explanatory tools such as class activation maps \citep{zhou_cam_2016}, were calculated for several inputs. Figure~\ref{fig:shap_values} shows SHAPs for Focus-s in MNIST, for CNN+Fcs in CIFAR-10, and for VGG+Fcs in CIFAR-10, demonstrating that the Focus-s network localized regions in MNIST characters, CNN+Fcs used local textures in CIFAR-10 instances whereas Vgg+Fcs was able to localize entire objects. The difference between SHAP outputs in separate networks shows that the ability to localize the entire object depends mostly on the power of the deep convolutional stacks (VGG) rather than the focusing layers.

\subsection{Other Data}
The next question was how the model would function when dealing with non-spatial or non-image inputs. To test this, the Boston House price regression, Reuters newswire classification and IMDB sentiment classification data sets (available in Keras \citep{keras}) were used, along with the DNA sequence (1D spatial) data set from the OpenML \citep{OpenML2013}. The setup and results were as follows.

\emph{Boston}: The house price data set includes 506 real house price values with 13 features. A two hidden-layer network with 64 neurons in the hidden layers was constructed. Both networks were trained with RmsProp \citep{keras} for 200 epochs with a batch size of 16. Table~\ref{tab:comp3} shows the minimum squared errors achieved over 10-fold cross-validation, and Figure~\ref{fig:losses}e shows training and validation errors from one of the runs. While the dense network achieved the minimum error, the mean error rates were almost identical and t-tests confirmed no statistical significance between the results (t=-0.47, p=0.64). 

\emph{Newswires}: There are 11,228 newswire instances in this data set. The input sequences of word indices were converted to vectors of 1000 elements (following the steps in Keras Reuters example in \citep{keras}). Then, a single hidden-layer network of 150 hidden neurons was constructed. The networks were trained for 30 epochs with a stochastic mini-batch gradient descent of batch size 64 and a learning rate of 0.01. The foci were initially spread out with $\sigma=0.25$. The average of the accuracies in 10 cross-validations presented in Table~\ref{tab:comp3} shows that the dense network performed marginally better than Focused-s (see Figure~\ref{fig:losses}e for training and validation errors from one of the runs). However, t-tests did not confirm statistical significance of the difference between the results (t=1.02, p=0.33).

\emph{DNA}: The problem in the DNA data set is to recognize particular junction codes in gene sequences. The data includes 3186 instances of 60 binary triples which encode nucleotides (A, C, G, T). Three classes indicate the middle of the sequence as one of the following, EI: exon-intron boundary; IE: intron-exon boundary; None: not a boundary. The OpenML description states that using the middle-right of the feature vector leads to more successful results, making this an appropriate challenge for focusing neurons. For this experiment, a network consisting of a single hidden layer with 60 hidden units was set up. The foci were initially spread out with $\sigma=0.025$. The networks were then trained for 200 epochs in 32 batches with a learning rate of 1e-3 for all parameters. Table~\ref{tab:comp3} shows that the focused network was marginally better than its dense counterpart, however the difference was not  confirmed as significant by t-test.

\emph{IMDB}: The data set (available in Keras~\citep{keras}) contains 25000 movie reviews labeled as positive/negative. In order not to repeat the Newswires test, the Fasttext architecture~\citep{joulin-etal-2017-bag}, based on bi-gram enhanced word input vectors, was employed. This architecture is composed of an embedding layer (400,50) followed by a global average pooling layer which computes a reduced hidden representation (50). Then, a single sigmoid neuron computes the output probability of positive/negative classes. The dense output neuron was replaced with a single focused neuron for the comparison. Although there was not much room for adaptive focusing, the neuron would nonetheless adjust the position of its receptive field. Thus the single neuron was initialized with the focus at the center $\mu=0.5$ with $\sigma=0.5$. The networks were trained for seven epochs (as in the Keras example) with the Adam optimizer \citep{adam}. Table~\ref{tab:comp3} indicates that the results were similar, with no statistical significance in the difference.

\begin{table}[]
	\caption{Test classification performance in other data sets. N (repeats)=5, T-tests results shown by t (p) \hl{* :significance}.}{
	\resizebox{\textwidth}{!}{%
		\begin{tabular}{@{}llll|lll|lll|lll@{}}
			\toprule
			& \multicolumn{3}{c}{Housing}                                   & \multicolumn{3}{c}{Newsgroups}                                     & \multicolumn{3}{c}{DNA}                                   & \multicolumn{3}{c}{IMDB}                                 \\ \midrule
			& Mn+std & \multicolumn{1}{c}{Min} & \multicolumn{1}{c}{t (p)} & Mn$\pm$std & \multicolumn{1}{c}{Max} & \multicolumn{1}{c}{t (p)} & Mn$\pm$std & \multicolumn{1}{c}{Max} & \multicolumn{1}{c}{t (p)} & Mn$\pm$std & \multicolumn{1}{c}{Max} & \multicolumn{1}{c}{t (p)} \\ \midrule
			
			Dense & \textbf{2.2$\pm$0.21}    & \textbf{1.77}                 & \multirow{2}{*}{2e-3 (0.99)}                      &\textbf{81.48$\pm$7e-3}    & \textbf{82.68}                & \multirow{2}{*}{1.02 (0.33)}                      & 95.2$\pm$0.01     & 96.23                 & \multirow{2}{*}{-1.5 (0.14)}                      &  \textbf{88.98$\pm$1e-4}    & 89.15                 & \multirow{2}{*}{0.06 (0.94)}                       \\ 
			
			Focus-s  &   2.21$\pm$0.26   & 1.93                  &                          &  81.05$\pm$6e-3&  82.1                     &                 &  \textbf{96.2$\pm$8e-3 }   & \textbf{97.02}                 &                        &  88.97$\pm$2e-3    &  \textbf{89.17}                  &                         \\ \bottomrule
		\end{tabular}}%
	}\label{tab:comp3}
\end{table}

\subsection{Training and test time}
In the synthetic random experiments, a training epoch of the focusing network took $\approx$1.3 times longer than the dense network. In MNIST, for the two hidden layer configuration, a single batch iteration of 512 instances took $\approx$24us for the focusing network and $\approx$21us for the dense network while training on an NVIDIA Tesla K40 GPU. A similar ratio was observed for CIFAR-10 (49us vs 42us). The overhead was due to the additional gradient computations and parameter updates. The testing time difference for 10000 inputs ($\approx$0.056s vs $\approx$0.053s) was negligible, as the only overhead was the calculation of the focus function, which does not change from input to input.

\section{Discussion}

The experiments demonstrated that focusing neurons can adapt and learn their local receptive field locations and sizes. This capacity enables the neurons to focus on informative features and steer away from the redundant inputs.

When placed after inputs in standard image recognition data sets, the focusing layers demonstrated significantly better performance than the dense layers. However, in relatively noisy and challenging data such as the MNIST-cluttered, CIFAR-10, and LFW-Faces the difference was remarkable. Moreover, the experiments showed that a focused network with 256 neurons in the hidden layers could work more effectively than a dense network with 256, 800 or 1024 neurons in the hidden layers. However, in 1D data sets the focusing network performed better than the dense network only in the (spatial) DNA data set.

The focusing neuron model is not designed for feature extraction. It is not translation invariant. It is not expected to compete with convolutional kernels which share weights for translation equivariant feature extraction. However, the experiments demonstrated that when used for classification of convolutional features, focusing layers may work similar to or in some cases better than dense layers (see CNN+Fcs results for MNIST-CLT and CIFAR-10). In transfer-learning, employed as a classification layer, the focusing layers used the flattened output of the convolution stacks as the input. It is perhaps, due to the highly refined feature representation or the reduced input length that no advantage could be observed over the dense layers.

Focusing neurons can establish local connections, but, their view is still a part of the whole input field. They contribute to the minimization of global errors by adapting to the local cues given by the local gradient. Though the focus parameters are only updated at the neuron level, they could collectively partition the whole input domain in many cases. In addition, they would form narrow or wider apertures according to different roles (feature extraction or classification).
However, adapting to the best input positions can cause neurons to become sensitive to the positional information and overfit. For example, in the MNIST-cluttered set, the manually tuned fixed focused model (Fixed-s) worked better than the trained model (Focused-s).


Nevertheless, from a biological perspective, it may be irrelevant to compare the different neuronal models because the brain has almost 10000 different types of nerve cells \citep{haykin1998,kandel_2006}. Brain networks are formed of many local clusters with dense and short-path connections and few long-range connections linking those clusters \citep{menon_2015}. Hence, the role of a focusing neuron can be different from the ones that are usually assigned to a fully connected neuron. For example, focusing neurons can be used to partition and distribute input spaces. To test this capacity, it is possible to design experiments in which the input is complex (multi-domain), for example, a concatenation of visual and spoken signals.

\subsection{Limitations of the Study}
The current implementation of the focusing neuron was one-dimensional because the initial aim was to produce a core module that is compatible with all settings/domains. One can speculate that a 2D implementation would perform better with 2D image inputs because it would take better advantage of the spatiality. In addition, the architectures tested in the experiments were shallow compared to the state-of-the-art deep networks such as residual networks \cite{resnet}. The next step will be to test the performance of focusing neurons in deeper networks on larger data sets, and on different tasks.

\section{Conclusion}
This paper has presented a new neuron model capable of learning its local receptive field region and size in the topological domain of its inputs. The new model comes with a differentiable receptive field, which is named here as a focus. Though our choice of Gaussian focus was aimed at creating locally connected receptive fields, other differentiable functions could also be selected. 

In synthetic and real data sets, the experiments demonstrated that focusing neurons can train their local receptive fields, thereby providing better generalization performance and sparser networks. Using focusing neuron layers as the top-level classifier in convolutional networks achieves better performance in some data sets such as MNIST-cluttered and CIFAR-10. However, as the convolutional stack gets deeper and features get refined (as in VGG-16), the advantage is lost. 

This work can be extended by exploring 2D-3D focusing neurons/layers, different focus control functions, recurrent neurons, dynamic focusing, applications in attention-based models, and modeling of multi-problem input domains.

\section*{Funding}
This work was supported by TUBITAK 1001 program (no:118E722), Isik University BAP program (no:16A202), and NVIDIA hardware donation of a Tesla K40 GPU unit.

\section*{Acknowledgments}
Thanks to Mr.\ İlker Çam for his contributions in the early development of the model. Thanks to Dr.\ Murat S. Ayhan and Dr.\ Olcay T. Yıldız for discussions on the model. Thanks to Dr.\ Mehmet Önal, Dr.\ Deniz Karlı for their comments on the weight initialization. Thanks to Dr.\ Emine Ekin and Dr.\ Robert J. Gray for reading and correcting the manuscript.
\section*{References}

\bibliography{focusv2}

\begin{thebibliography}{10}
\expandafter\ifx\csname url\endcsname\relax
  \def\url#1{\texttt{#1}}\fi
\expandafter\ifx\csname urlprefix\endcsname\relax\def\urlprefix{URL }\fi
\expandafter\ifx\csname href\endcsname\relax
  \def\href#1#2{#2} \def\path#1{#1}\fi

\bibitem{charles2009}
C.~D. Gilbert, W.~Li, V.~Piech, Perceptual learning and adult cortical
  plasticity, The Journal of Physiology 30 (2009) 2743–2751.

\bibitem{merzenich2014}
M.~M. Merzenich, T.~M.~V. Vleet, M.~Nahum, Brain plasticity-based therapeutics,
  Frontiers in Human Neuroscience 8 8 (2014) 335.

\bibitem{power2010}
J.~D. Power, D.~A. Fair, B.~L. Schlaggar, S.~E. Petersen, The development of
  human functional brain networks, Neuron 67 (2010) 735–748.

\bibitem{menon_2015}
V.~Menon, Large-Scale Functional Brain Organization, Vol.~2, Elsevier, 2015,
  pp. 449--459.
\newblock \href {http://dx.doi.org/10.1016/j.tics.2011.08.003}
  {\path{doi:10.1016/j.tics.2011.08.003}}.

\bibitem{bartunov2018}
S.~Bartunov, A.~Santoro, B.~A. Richards, G.~E. Hinton, T.~Lillicrap, Assessing
  the scalability of biologically-motivated deep learning algorithms and
  architectures, in: Advances in Neural Information Processing Systems.
\newblock \href {http://dx.doi.org/10.1016/B978-0-12-397025-1.00024-5}
  {\path{doi:10.1016/B978-0-12-397025-1.00024-5}}.

\bibitem{szegedy2016}
C.~Szegedy, S.~Ioffe, V.~Vanhoucke, A.~Alemi, Inception-v4, inception-resnet
  and the impact of residual connections on learning, in: Thirty-First AAAI
  Conference on Artificial Intelligence, 2017, 1602.07261v2.

\bibitem{larsson2017}
G.~Larsson, M.~Maire, G.~Shakhnarovich, Fractalnet: Ultra-deep neural networks
  without residuals, in: Int. Conf. on Learning Representations, 2017.

\bibitem{urban2017}
G.~Urban, Do deep convolutional nets really need to be deep and convolutional?,
  in: Int. Conf. on Learning Representations, 2017.

\bibitem{srivastava2015}
R.~K. Srivastava, K.~Greff, J.~Schmidhuber, Highway networks, in: Int. Conf. on
  Machine Learning Deep Learning workshop, 2015.

\bibitem{xu2015}
K.~Xu, J.~L. Ba, R.~K. et~al., Show, attend and tell: Neural image caption
  generation with visual attention, in: Int. Conf. on Machine Learning,
  Vol.~37, 2015, pp. 2048--2057.

\bibitem{ba2014}
J.~Ba, V.~Mnih, K.~Kavukcuoglu, Multiple object recognition with visual
  attention, CoRR abs/1412.7755.

\bibitem{Vinyals2015}
O.~{Vinyals}, A.~{Toshev}, S.~{Bengio}, D.~{Erhan}, Show and tell: A neural
  image caption generator, in: IEEE Conf. on Computer Vision and Pattern
  Recognition (CVPR), 2015, pp. 3156--3164.
\newblock \href {http://dx.doi.org/10.1109/CVPR.2015.729893}
  {\path{doi:10.1109/CVPR.2015.729893}}.

\bibitem{Floreano2008}
D.~Floreano, P.~D{\"u}rr, C.~Mattiussi, Neuroevolution: from architectures to
  learning, Evolutionary Intelligence 1~(1) (2008) 47--62.
\newblock \href {http://dx.doi.org/10.1007/s12065-007-0002-4}
  {\path{doi:10.1007/s12065-007-0002-4}}.

\bibitem{soltoggio_2018}
A.~Soltoggio, K.~O. Stanley, S.~Risi, Born to learn: the inspiration, progress,
  and future of evolved plastic artificial neural networks, Neural Networks 108
  (2018) 48--67.
\newblock \href {http://dx.doi.org/10.1016/j.neunet.2018.07.013}
  {\path{doi:10.1016/j.neunet.2018.07.013}}.

\bibitem{romero2015}
A.~Romero, N.~Ballas, S.~E. Kahou, A.~Chassang, C.~Gatta, Y.~Bengio, Fitnets:
  Hints for thin deep nets, in: Int. Conf. on Learning Representations, 2015.

\bibitem{baker2017}
B.~Baker, O.~Gupta, N.~Naik, R.~Raskar, Designing neural network architectures
  using reinforcement learning, in: Int. Conf. on Learning Representations,
  2017.

\bibitem{liu2018}
H.~Liu, K.~Simonyan, Y.~Yang, Darts: Differentiable architecture search, in:
  Int. Conf. on Learning Representations, 2019.

\bibitem{coates2011}
A.~Coates, A.~Y. Ng, Selecting receptive fields in deep networks, in: Advances
  in Neural Information Processing Systems, 2011.

\bibitem{fiesler1994}
E.~Fiesler, Comparative bibliography of ontogenic neural networks, in: Int.
  Conf. on Artificial Neural Networks, Springer, 1994.

\bibitem{hassibi93}
B.~Hassibi, D.~G. Stork, G.~J. Wolff, Optimal brain surgeon and general network
  pruning, in: IEEE Int. Conf. on Neural Networks, Vol.~1, 1993, pp. 293--299.

\bibitem{han2015}
S.~Han, J.~Pool, J.~Tran, W.~J. Dally, Learning both weights and connections
  for efficient neural networks, in: Advances in Neural Information Processing
  Systems, 2015, pp. 1135--1143.

\bibitem{cortes2017}
C.~Cortes, X.~Gonzalvo, V.~Kuznetsov, M.~Mohri, S.~Yang, {A}da{N}et: Adaptive
  structural learning of artificial neural networks, in: Int. Conf. on Machine
  Learning, 2017, pp. 874--883.

\bibitem{Serre_2007}
T.~Serre, L.~Wolf, S.~Bileschi, M.~Riesenhuber, T.~Poggio, Robust object
  recognition with cortex-like mechanisms, IEEE Pattern Analysis and Machine
  Intelligence 29 (2007) 411--26.
\newblock \href {http://dx.doi.org/10.1109/TPAMI.2007.56}
  {\path{doi:10.1109/TPAMI.2007.56}}.

\bibitem{Masquelier_2007}
T.~Masquelier, T.~Serrea, S.~Thorpe, T.~Poggio, Learning simple and complex
  cells-like receptive fields from natural images: a plausibility proof,
  Journal of Vision 7 (2007) 81.

\bibitem{olshausen_1996}
B.~A. Olshausen, D.~J. Field, Emergence of simple-cell receptive field
  properties by learning a sparse code for natural images, Nature 381 (1996)
  607–609.

\bibitem{cam2017}
I.~\c{C}am, F.~B. Tek, Odaklanan n\"{o}ron (focusing neuron), in: 25th Signal
  Processing and Communications Applications Conference (SIU), 2017, pp. 1--4.
\newblock \href {http://dx.doi.org/10.1109/SIU.2017.7960632}
  {\path{doi:10.1109/SIU.2017.7960632}}.

\bibitem{tek_2019}
F.~B. Tek, Uyarlanır yerel bağlı nöron modelinin İncelemesi, Bilişim
  Teknolojileri Dergisi 12 (2019) 307--317.

\bibitem{stoeckli2017}
E.~Stoeckli, Where does axon guidance lead us?, F1000Research 6~(78).

\bibitem{tracey2017}
T.~A. Suter, Z.~J. DeLoughery, A.~Jaworski, Meninges-derived cues control axon
  guidance, Developmental Biology 430 (2017) 1--10.

\bibitem{lecun1998a}
Y.~LeCun, L.~Bottou, Y.~Bengio, P.~Haffner, Gradient-based learning applied to
  document recognition, in: Proc. of the IEEE, Vol.~86, 1998, pp. 2278--2324.

\bibitem{jaderberg2015}
M.~Jaderberg, K.~Simonyan, A.~Zisserman, K.~Kavukcuoglu, Spatial transformer
  networks, in: Advances in Neural Information Processing Systems, Vol.~28,
  2015.

\bibitem{xiao2017}
H.~Xiao, K.~Rasul, R.~Vollgraf, Fashion-mnist: a novel image dataset for
  benchmarking machine learning algorithms, arXiv cs.LG/1708.07747.

\bibitem{cifar10}
A.~Krizhevsky, Learning multiple layers of features from tiny images, Tech.
  rep., Canadian Institute For Advanced Research (2009).

\bibitem{lfw}
G.~B. Huang, M.~Ramesh, T.~Berg, E.~Learned-Miller, Labeled faces in the wild:
  A database for studying face recognition in unconstrained environments, Tech.
  Rep. 07-49, University of Massachusetts, Amherst (Oct 2007).

\bibitem{keras}
F.~Chollet, et~al., Keras, \url{https://keras.io} (2015).

\bibitem{OpenML2013}
J.~Vanschoren, J.~N. van Rijn, B.~Bischl, L.~Torgo, Openml: Networked science
  in machine learning, SIGKDD Explorations 15 (2013) 49--60.

\bibitem{Hubel_1962}
D.~. T.~W. Hubel, Receptive fields, binocular interaction and functional
  architecture in the cat’s visual cortex., J Physiology 160 (1962) 106--54.

\bibitem{Chang_2017}
L.~Chang, D.~Y.Tsao, The code for facial identity in the primate brain, Cell
  169 (2017) 1013--1028.

\bibitem{Poggio_2013}
T.~Poggio, T.~Serre, {M}odels of visual cortex, Scholarpedia 8~(4) (2013) 3516.
\newblock \href {http://dx.doi.org/10.4249/scholarpedia.3516}
  {\path{doi:10.4249/scholarpedia.3516}}.

\bibitem{Hebb}
D.~O. Hebb, The Organization of Behaviour, John Wiley \& Sons, 1949.

\bibitem{baldi_2016}
P.~Baldi, P.~Sadowski, A theory of local learning, the learning channel, and
  the optimality of backpropagation, Neural Networks 83 (2016) 51--74.

\bibitem{Rosenblatt}
F.~Rosenblatt, The perceptron: A probabilistic model for information storage
  and organization in the brain, cornell aeronautical laboratory, Psychological
  Review 65~(6) (1958) 386–408.

\bibitem{Minsky_1969}
M.~Minsky, S.~Papert, Perceptrons, MIT Press., Cambridge, MA, 1969.

\bibitem{fukushima_1983}
K.~Fukushima, S.~Miyake, T.~Ito, Neocognitron: A neural network model for a
  mechanism of visual pattern recognition, IEEE Trans. Systems, Man, and
  Cybernetics 13 (1983) 826--834.

\bibitem{haykin1998}
S.~Haykin, Neural Networks: A Comprehensive Foundation, 2nd Edition, Prentice
  Hall PTR, Upper Saddle River, NJ, USA, 1998.

\bibitem{hagan2014}
M.~T. Hagan, H.~B. Demuth, M.~H. Beale, Neural Network Design, Martin Hagan,
  2014.

\bibitem{gitcode}
F.~B. Tek (2018).
\newblock \href{https://github.com/btekgit/FocusingNeuron.git}{[link]}.
\newline\urlprefix\url{https://github.com/btekgit/FocusingNeuron.git}

\bibitem{lecun1990}
Y.~LeCun, J.~S. Denker, S.~A. Solla, Optimal brain damage, in: D.~S. Touretzky
  (Ed.), Advances in Neural Information Processing Systems, Morgan-Kaufmann,
  1990, pp. 598--605.

\bibitem{elizondo1997}
D.~Elizondo, R.~Fiesler, A survey of partially connected neural networks., Int
  J. Neural Systems 8 (1997) 535--568.

\bibitem{howard2017}
A.~G. Howard, M.~Zhu, B.~Chen, D.~Kalenichenko, W.~Wang, T.~Weyand,
  M.~Andreetto, H.~Adam, Mobilenets: Efficient convolutional neural networks
  for mobile vision applications (2017).
\newblock \href {http://arxiv.org/abs/1704.04861} {\path{arXiv:1704.04861}}.

\bibitem{han2015deep}
S.~Han, H.~Mao, W.~J. Dally, Deep compression: Compressing deep neural networks
  with pruning, trained quantization and huffman coding, in: Int. Conf. on
  Learning Representations, 2015.

\bibitem{Manessi2017}
F.~Manessi, A.~Rozza, S.~Bianco, P.~Napoletano, R.~Schettini, Automated pruning
  for deep neural network compression, IEEE, 2018.
\newblock \href {http://dx.doi.org/10.1109/icpr.2018.8546129}
  {\path{doi:10.1109/icpr.2018.8546129}}.

\bibitem{wu2016quantized}
J.~Wu, C.~Leng, Y.~Wang, Q.~Hu, J.~Cheng, Quantized convolutional neural
  networks for mobile devices, in: IEEE Conf. on Computer Vision and Pattern
  Recognition, 2016, pp. 4820--4828.

\bibitem{goodfellow2016}
I.~GoodFellow, Y.~Bengio, A.~Courville, Deep Learning, The MIT Press, 2016.

\bibitem{kung1988}
S.~Y. Kung, J.~N. Hwang, S.~W. Sun, Efficient modeling for multilayer
  feed-forward neural nets, in: Int. Conf. on Acoustics, Speech, and Signal
  Proc., Vol.~4, 1988, pp. 2160--2163.

\bibitem{lecun_1989_local}
Y.~LeCun, Generalization and network design strategies., Tech. Rep.
  CRG-TR-89-4, University of Toronto (1989).

\bibitem{taigman2014}
Y.~Taigman, M.~Yang, M.~Ranzato, L.~Wolf, Deepface: Closing the gap to
  human-level performance in face verification, 2014 IEEE Conf. on Computer
  Vision and Pattern Recognition (2014) 1701--1708.

\bibitem{rowley1998}
H.~A. Rowley, S.~Baluja, T.~Kanade, Neural network-based face detection, IEEE
  Trans. Pattern Analysis Machine Intelligence 20 (1998) 23--38.

\bibitem{gregor2010}
K.~Gregor, Y.~LeCun, Emergence of complex-like cells in a temporal product
  network with local receptive fields, arXiv abs/1006.0448.

\bibitem{munder_2006}
S.~{Munder}, D.~M. {Gavrila}, An experimental study on pedestrian
  classification, IEEE Trans. on Pattern Analysis and Machine Intelligence
  28~(11) (2006) 1863--1868.

\bibitem{pang2017}
L.~Pang, Y.~Lan, J.~Xu, J.~Guo, X.~Cheng, Locally smoothed neural networks, in:
  Proc. Machine Learning Research 77, 2017.

\bibitem{cam_2018}
I.~Cam, F.~B. Tek, Learning filter scale and orientation in cnns, arXiv
  preprint arXiv:1803.00388.

\bibitem{rbf}
M.~J.~L. Orr, Introduction to radial basis function networks (1996).

\bibitem{Kohonen_lvq_1995}
T.~Kohonen, Learning vector quantization, in: M.~Arbib (Ed.), The Handbook of
  Brain Theory and Neural Networks, MIT Press, 1995.

\bibitem{kohonen1990}
T.~Kohonen, The self-organizing map, Proceedings of the IEEE 78~(9) (1990)
  1464--1480.

\bibitem{esposito_2016}
U.~Esposito, Investigating connectivity in brain-like networks, Ph.D. thesis,
  The University of Sheffield, UK (2016).

\bibitem{bodenhausen_1990}
U.~Bodenhausen, A.~Waibel, The tempo 2 algorithm: Adjusting time-delays by
  supervised learning, in: Advances in Neural Information Processing Systems,
  1990, pp. 155--161.

\bibitem{Gerstner_2002}
W.~Gerstner, W.~M. Kistler, Mathematical formulations of {H}ebbian learning,
  Biological Cybernetics 87~(404–415).
\newblock \href {http://dx.doi.org/10.1007/s00422-002-0353-y}
  {\path{doi:10.1007/s00422-002-0353-y}}.

\bibitem{triesch_2007}
J.~Triesch, Synergies between intrinsic and synaptic plasticity mechanisms.,
  Neural Computing 19 (2007) 885–909.
\newblock \href {http://dx.doi.org/10.1162/neco.2007.19.4.885}
  {\path{doi:10.1162/neco.2007.19.4.885}}.

\bibitem{Oja1982}
E.~Oja, Simplified neuron model as a principal component analyzer, Journal of
  Mathematical Biology 15~(3) (1982) 267--273.

\bibitem{miconi_2018}
T.~Miconi, J.~Clune, K.~O. Stanley, Differentiable plasticity: training plastic
  networks with gradient descent, in: Int. Conf. on Machine Learning, 2018.

\bibitem{huang_2015}
G.~B. Huang, Z.~Bai., L.~Kasun, C.~Vong, Local receptive fields based extreme
  learning machine., IEEE Computational Intelligence Magazine 10 (2015) 18--29.

\bibitem{titi1998}
L.~Itti, C.~Koch, E.~Niebur, A model of saliency-based visual attention for
  rapid scene analysis, IEEE Trans. on Pattern Analysis and Machine
  Intelligence 20 (1998) 1254--1259.

\bibitem{Olshausen4700}
B.~Olshausen, C.~Anderson, D.~Van~Essen, A neurobiological model of visual
  attention and invariant pattern recognition based on dynamic routing of
  information, Journal of Neuroscience 13~(11) (1993) 4700--4719.

\bibitem{CheungWO16}
B.~Cheung, E.~Weiss, B.~A. Olshausen, Emergence of foveal image sampling from
  learning to attend in visual scenes, in: Int. Conf. on Learning
  Representations, 2017.

\bibitem{sabour2017}
S.~Sabour, N.~Frosst, G.~E. Hinton, Dynamic routing between capsules, in:
  Advances in Neural Information Processing Systems, 2017.

\bibitem{theano}
{Theano Dev. Team}, {Theano: A {Python} framework for fast computation of
  mathematical expressions} (May 2016).

\bibitem{tensorflow}
M.~Abadi, P.~Barham, Tensorflow: A system for large-scale machine learning, in:
  12th USENIX Symposium on Operating Systems Design and Implementation, 2016,
  pp. 265--283.

\bibitem{Lindeberg2011}
T.~Lindeberg, Generalized {G}aussian scale-space axiomatics comprising linear
  scale-space, affine scale-space and spatio-temporal scale-space, Journal of
  Mathematical Imaging and Vision 40~(1) (2011) 36--81.

\bibitem{he2015}
K.~He, X.~Zhang, S.~Ren, J.~Sun, Delving deep into rectifiers: Surpassing
  human-level performance on imagenet classification, in: IEEE Int. Conf. on
  Computer Vision, 2015, pp. 1026--1034.

\bibitem{glorot2010}
X.~Glorot, Y.~Bengio, Understanding the difficulty of training deep feedforward
  neural networks, in: Proc. of Machine Learning Research, Vol.~9, 2010, pp.
  249--256.

\bibitem{fsdd}
Z.~Jackson, C.~Souza, J.~Flaks, Y.~Pan, H.~Nicolas, A.~Thite,
  \href{https://github.com/Jakobovski/free-spoken-digit-dataset}{Free spoken
  digits dataset} (Aug 2018).
\newblock \href {http://dx.doi.org/10.5281/zenodo.1342401}
  {\path{doi:10.5281/zenodo.1342401}}.
\newline\urlprefix\url{https://github.com/Jakobovski/free-spoken-digit-dataset}

\bibitem{blundel_2015}
C.~Blundell, J.~Cornebise, K.~Kavukcuoglu, D.~Wierstra, Weight uncertainty in
  neural networks, in: Int. Conf. on Machine Learning, 2015, pp. 1613--1622.

\bibitem{simonyan2014deep}
K.~Simonyan, A.~Zisserman, Very deep convolutional networks for large-scale
  image recognition (2014).
\newblock \href {http://arxiv.org/abs/1409.1556} {\path{arXiv:1409.1556}}.

\bibitem{resnet}
K.~He, X.~Zhang, S.~Ren, J.~Sun, Deep residual learning for image recognition,
  in: IEEE Conf. on Computer Vision and Pattern Recognition, 2015.

\bibitem{lundberg_2017}
S.~Lundberg, S.-I. Lee, A unified approach to interpreting model predictions,
  in: Advances in Neural Information Processing Systems, 2017.

\bibitem{zhou_cam_2016}
B.~{Zhou}, A.~{Khosla}, A.~{Lapedriza}, A.~{Oliva}, A.~{Torralba}, Learning
  deep features for discriminative localization, in: IEEE Conf. on Computer
  Vision and Pattern Recognition, 2016, pp. 2921--2929.

\bibitem{joulin-etal-2017-bag}
A.~Joulin, E.~Grave, P.~Bojanowski, T.~Mikolov, Bag of tricks for efficient
  text classification, in: Conf. of the European Chapter of the Association for
  Computational Linguistics: Vol 2, Short Papers, 2017, pp. 427--431.

\bibitem{adam}
D.~Kingma, J.~Ba, Adam: A method for stochastic optimization, in: Int. Conf. on
  Learning Representations, 2014.

\bibitem{kandel_2006}
E.~R. Kandel, In search of memory: The emergence of a New Science of Mind, W.
  W. Norton \& Company, 2006.

\bibitem{lecun2012}
Y.~LeCun, L.~Bottou, G.~B. Orr, K.-R. M{\"u}ller, Efficient BackProp, Springer,
  Heidelberg, 2012, pp. 9--48.

\bibitem{kumar2017}
S.~K. Kumar, On weight initialization in deep neural networks, arXiv
  abs/1704.08863.

\end{thebibliography}

\appendix

\section{Weight Initialization}\label{apx_w_init}
An important aspect of the weight initialization is to sustain the variance of the signals that propagate through the layers \citep{he2015,glorot2010,lecun2012,kumar2017}. Hence, the objective is to have the variance of the output $y$ equal to the variance of the input $x_i$, $\mathtt{Var}(y)\!=\!\mathtt{Var}(x)$. Kumar's \citep{kumar2017} approach and notation can be used to derive an appropriate weight initialization scheme for the focusing neuron model. Here, the activation-transfer function can be omitted, because the focusing model has no extra effect on it. Assume $x_i$ and $w_i$ are both independent and identically distributed (i.i.d) variables and $\phi(\tau(i,\theta))$ (shortly $\phi(i)$) is a deterministic function of $i$. The weights will be identically sampled from a zero mean distribution; hence the expected value is zero $\mathbb{E}\left[w_i\right]=0$. However, a second initialization scheme is possible if non-identical distributions are used. Let us start by writing the variance of the output $y$ in terms of the weights, inputs, and $\phi(i)$.
\begin{gather}
y = \sum_{i=1}^{m} w_{i}\phi(i) x_{i} + b  \\
\mathtt{Var}(y) = \mathbb{E}\left[ y^2 \right]- \mathbb{E}^2\left[ y \right]\\
\mathtt{Var}(y) = \mathbb{E}\left[ \Big(\sum_{i=1}^{m} w_{i}\phi(i) x_{i}\Big)^2 \right]- \mathbb{E}^2\left[ \sum_{i=1}^{m} w_{i}\phi(i) x_{i} \right]
\label{init1}
\end{gather} 
Since $x_i$ and $w_i$ are independent and  $\mathbb{E}\left[w_i\right]=0$, the second term on the right reduces to 0. If the first term is examined,
\begin{gather}
\mathtt{Var}(y) = \mathbb{E}\left[ \big(\sum_{i=1}^{m} w_{i}\phi(i) x_{i}\big)^2 \right] \\
\mathtt{Var}(y) = \sum_{i=1}^{m} \phi^2(i) \mathbb{E}\left[w_{i}^2\right] \mathbb{E}\left[x_{i}^2\right] + \label{init2} \\ 
2\sum_{i=1}^{m}\sum_{k=i+1}^{m} \phi(i) \phi(k) \mathbb{E}\left[ x_{i}x_{k}w_{i}w_{k}\right] \notag
\end{gather} 
Again since $x_i$, $w_i$, $x_k$, $w_k$, are all independent and since $\mathbb{E}\left[w_i\right]=0$, the double summation in (\ref{init2}) reduces to 0. Therefore,
\begin{equation}
\mathtt{Var}(y)= \sum_{i=1}^{m} \phi^2(i) \mathbb{E}\left[w_{i}^2\right] \mathbb{E}\left[x_{i}^2\right]
\label{init3}
\end{equation}
Now, the notation can be simplified by writing $s^2_{w_i}$, $s^2_{x_i}$, $s^2_y$ for the variances of the weight $i$, input $i$, and output $y$, respectively. In addition, $\mu_{x_i}$ denotes the mean for input $i$. Remember that weights are sampled from zero mean $\mu_{w_i}=0$. Let us rewrite (\ref{init3}) as
\begin{equation}
s^2_y= \sum_{i=1}^{m} \phi^2(i) (s^2_{w_i}) (s^2_{x_i}+\mu^2_{x_i})
\label{yvar}
\end{equation}
Now, set constant variance for $y$ (e.g., $s^2_y=1$) to solve the weight variance in two different ways. Remember that $x_i$ are identical, so $\mu_{x_i}=\mu_{x}$ and $s_{x_i}=s_{x}$. First, assume identical distribution and equal variance for the weights (i.e., $s^2_{w_i}=s^2_{w_k}=s^2_{w}$). Thus, the variance of each weight can be expressed as:
\begin{equation}
s^2_{w}=\frac{1}{(s^2_{x}+\mu^2_{x})\sum_{i=1}^{m} \phi^2(i)}
\label{init_sum1}
\end{equation}
If we set $s^2_{x}=1$ and $\mu^2_{x}=0$, $s^2_{w_i}$ reduces to one over the squared norm of the $\phi$ coefficient vector. 
However, instead of equal variance weights, one can sample the weights non-identically from different distributions to create equal variance in the products $w_i*\phi(i)$. For example, if each term in the summation (\ref{yvar}) receives $1/m$ variance, for a total variance of 1, it is possible to initialize the independent weights with the following variance:
\begin{equation}
s^2_{w_i}=\frac{1}{m(s^2_{x}+\mu^2_{x})\phi^2(i)}
\label{init_indiv}
\end{equation}
The different transfer functions require different scalers to provide the expected output variance \citep{he2015,kumar2017}. The current implementation used (\ref{init_sum1}) and sampled the weights uniformly with $U\left[-\sqrt{6}/s_{w},\sqrt{6}/s_{w}\right]$ for the RELU activations.

\end{document}